\documentclass{article} % For LaTeX2e
\usepackage{booktabs}
\usepackage{iclr2026_conference,times}

\usepackage{amsmath,amsfonts,bm}

\def\eqref#1{equation~\ref{#1}}
\def\1{\bm{1}}
\DeclareMathAlphabet{\mathsfit}{\encodingdefault}{\sfdefault}{m}{sl}
\SetMathAlphabet{\mathsfit}{bold}{\encodingdefault}{\sfdefault}{bx}{n}
\let\introtakeawaybox\relax
\newcommand{\introtakeawaybox}[2][]{%
  \begin{tcolorbox}[%
    colback=ThemeOrange!10,
    colframe=ThemeGoldBrown!90,
    boxrule=0.5pt,
    arc=0mm,
    left=5pt,
    right=5pt,
    top=2pt,
    bottom=2pt,
    fontupper={\fontsize{9.5pt}{11.75pt}\selectfont}
  ]
    \IfNoValueTF{#1}{\textbf{Contributions:} #2}{\textbf{Contributions #1:} #2}%
  \end{tcolorbox}%
  \vspace*{-0.1cm}
}

\usepackage{url}
\usepackage{booktabs}
\usepackage{amsfonts}
\usepackage{nicefrac}
\usepackage{microtype}
\usepackage{amssymb}
\usepackage{mathtools}
\usepackage{array}
\usepackage{caption}
\usepackage{minitoc}
\usepackage{amsthm}
\usepackage{wrapfig}
\usepackage{fontawesome5}

\usepackage{setspace}
\usepackage{tabularx}
\usepackage{multirow}
\usepackage{makecell}
\usepackage{placeins}
\usepackage{svg}
\usepackage{longtable}
\usepackage{xspace}
\usepackage{microtype}
\usepackage{tikz}
\usepackage{tcolorbox}
\usetikzlibrary{positioning}
\usepackage{algpseudocode}
\usepackage{lineno}
\usepackage{pifont}
\usepackage{bbding}
\usepackage{array}
\usepackage{pifont}
\usepackage{amsmath}
\usepackage{caption}
\usepackage{subcaption}
\usepackage{colortbl}
\usepackage{varwidth}
\usepackage{lipsum}
\usepackage{floatrow}
\usepackage{minitoc}
\usepackage[inline]{enumitem}

\usepackage{xcolor,colortbl}
\definecolor{Gray}{gray}{0.94}

\usepackage[subfigure]{tocloft}
\usepackage[toc,page,header]{appendix}
\usepackage{minitoc}

\usepackage[ruled,vlined,linesnumbered]{algorithm2e}
\usepackage{varwidth}
\usepackage{graphicx}
\usepackage{amsmath, amssymb}
\usepackage{newtxtext,newtxmath}
\usepackage[export]{adjustbox}
\usepackage{setspace} % Appx spacing

\usepackage{newtxmath} % More beautiful math (also used in LIMA)
\usepackage[font=small,labelfont=it]{caption} % longtable
\newcommand{\ours}{\textsc{ADAPT}\xspace}

\usepackage[pagebackref=false,breaklinks=true,colorlinks,bookmarks=false]{hyperref}
\usepackage{cleveref}
\hypersetup{linkcolor=[rgb]{0.7,0.1,0.1}}
\hypersetup{citecolor=[rgb]{0.4,0.15,0.95}}
\definecolor{cvprblue}{rgb}{0.21,0.49,0.74}
\hypersetup{urlcolor  = cvprblue}

\definecolor{olmoeDarkYellow}{HTML}{fdac15}
\definecolor{defaultblue}{HTML}{0077B6}
\definecolor{defaultlightblue}{HTML}{00B4D8}
\definecolor{blue}{HTML}{03045E}
\definecolor{blueb}{HTML}{0077B6}
\definecolor{bluec}{HTML}{00B4D8}
\definecolor{blued}{HTML}{90E0EF}
\definecolor{bluee}{HTML}{CAF0F8}

\newcommand{\flan}{\textsc{Flan V2}}
\newcommand{\cott}{\textsc{CoT}}
\newcommand{\dolly}{\textsc{Dolly}}
\newcommand{\bbh}{\textsc{BBH}}

\newcommand{\oasst}{\textsc{Open Assistant 1}}

\newcommand{\mmlu}{\textsc{MMLU}}

\newcommand{\llamasmall}{\textsc{Llama-2-7B}}

\newcommand{\regmix}{RegMix\xspace}

\title{Rethinking Data Curation in LLM Training: \\Online Reweighting  Offers Better Generalization than Offline Methods}

\newcommand\blfootnote[1]{%
  \begingroup
  \renewcommand\thefootnote{}\footnote{#1}%
  \addtocounter{footnote}{-1}%
  \endgroup
}

\author{\parbox{13cm}{Wanru Zhao\textsuperscript{1}\thanks{Corresponding to: Wanru Zhao (wz341@cam.ac.uk)} , Yihong Chen\textsuperscript{2}, Yuzhi Tang\textsuperscript{3 $\dagger$}, Wentao Ma\textsuperscript{3 $\dagger$}, Shengchao Hu \textsuperscript{4}, \\Shell Xu Hu\textsuperscript{5}, Alex Iacob\textsuperscript{1}, Abhinav Mehrotra\textsuperscript{5}, Nicholas D.\ Lane \textsuperscript{1}} \vspace{1.5mm}\\
\textsuperscript{1} University of Cambridge, \quad
\textsuperscript{2} OATML, University of Oxford, \\
\textsuperscript{3} University of Toronto, \quad
\textsuperscript{4} Shanghai Jiao Tong University, \quad
\textsuperscript{5} Samsung AI Center 
\blfootnote{$\dagger$ Equal contribution.}
}
\iclrfinalcopy % Uncomment for camera-ready version, but NOT for submission.
\begin{document}

\maketitle

\blfootnote{$\dagger$ Equal contribution.}
\addtocontents{toc}{\protect\setcounter{tocdepth}{-1}}

\begin{abstract}
Data curation is a critical yet underexplored area in large language model (LLM) training. Existing methods, such as data selection and mixing, operate in an \textit{offline} paradigm, detaching themselves from training. This separation introduces engineering overhead and makes the curation brittle: the entire pipeline must be re-run under model/task shifts. Moreover, offline methods alter data size through hard filtering or resampling, often sacrificing data diversity and harming generalization.

%a new line to highlight rethinking
We propose to rethink data curation as an \textit{online} reweighting problem, where sample importance is dynamically adjusted during training via loss weighting rather than static pre-processing. 
Specifically, we introduce \textbf{ADAPT} (\textbf{A}daptive \textbf{D}ata reweighting for \textbf{P}retraining and Fine\textbf{T}uning), a dynamic online framework that reweights training samples with adaptive per-sample learning rates guided by similarity-based quality signals, without changing the number of training samples. 
Unlike offline methods that enforce a static data distribution, \ours{} acts as an implicit curriculum learner, progressively shifting focus from coarse-grained patterns to fine-grained semantic distinctions as the model evolves.
%This view preserves data diversity, adapts continuously to evolving model states, and yields a better performance–FLOPs tradeoff.
%
%\ours{} integrates reweighting directly into the optimization loop with minimal overhead. 
Experiments on both instruction tuning and large-scale pretraining show that \ours{} consistently outperforms offline selection/mixing and prior online methods, achieving stronger cross-benchmark generalization under equal FLOPs.\\
\includegraphics[height=2.5ex]{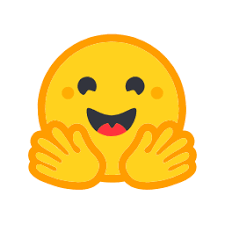} \url{https://huggingface.co/DATA-ADAPT} \\
\includegraphics[height=2.5ex]{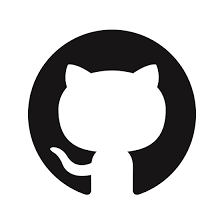} \url{https://github.com/Ryan0v0/ADAPT}
\end{abstract}

\section{Introduction}

Large language models (LLMs) have demonstrated remarkable capabilities across diverse tasks
\citep{brown2020language, chowdhery2022palm, touvron2023llama}, 
yet their generalization usually hinges on the quality of the training data, including its diversity and mixing 
\citep{hoffmann2022training, kaplan2020scaling}. While some recent efforts seek to circumvent the data quality issue and improve generalization through architectural interventions~\citep{peyrard2022invariant,chen2023improving,bengio2025superintelligent}, most pragmatic practices still rely on curating the ``right'' dataset for the model and the task.
% For foundation models, generalization is models must not only fit training distributions but also transfer across domains and benchmarks.
% A central ingredient to state-of-the-art open LLMs  is the large amount of high-quality data that these models are trained on. 
% Getting the right dataset and data mixture is painful and any LLM developer has to go through the laborious, time-consuming, energy-intensive and expensive steps of processing and filtering this crude data.
Data curation -- deciding what data to keep and how to weight it -- has thus become a cornerstone of large language model development. 
On the other hand, the process of data curation has largely remained a secret sauce with pipelines relying on opaque heuristics, ad-hoc engineering choices, or costly trial-and-error, rather than principled frameworks. 

% To give a taste of how challenging data curation can be, we refer readers to several works on language model research in different branches, as stated in Section~\ref{sec:related_work}.

% balancing, mixing, first, say that there are a lot of work and we comb through the piles of terminology, 
%second two main camps as follows
% Existing work on data curation, including \emph{data selection} and \emph{data mixing}, identifies and retains a subset of ``valuable'' examples, aiming to reduce compute while preserving performance. 
Current approaches to data curation largely follow an \textit{offline} paradigm, falling into two camps: data selection, which keeps a subset of valuable examples, and data mixing, which adjusts sampling frequencies to rebalance distributions.
% Data selection identifies and retains only the most valuable examples, reducing computational costs but potentially discarding useful information. 
% Data reweighting assigns fractional importance to examples, offering finer granularity. % but facing scalability challenges in large-scale training. 
Despite their differences, both approaches typically follow a multi-stage pipeline: (1) train or extract features with a proxy model, (2) compute quality signals as selection or weighting criteria on a validation set, and (3) retrain the main model on the curated data. 
%
% For example, pretraining dataset curation typically involves many stages including deduplication, various ad-hoc filters to remove noisy and undesirable content, and heuristic or intuition-based reweighting of sources. % Such curation is often applied to large-scale web scrapes. 
% 
% Or: pretraining dataset curation pipelines typically involve a series of isolated stages, each of which may be heuristic or ad-hoc. 
While effective in some cases, this paradigm suffers from fundamental drawbacks~(Sec.~\ref{sec:revisit}) due to its static and subtractive nature. First, it ignores training dynamics: the value of a data sample is not static but changes as the model learns. Offline selection freezes this value based on a proxy model, creating a misalignment with the training model's evolving needs. Second, it compromises diversity: by relying on hard filtering to select a fixed subset, offline methods risk discarding the broad data distribution essential for robust generalization. Consequently, the resulting subsets often overfit to specific benchmarks and are brittle, which requires the entire pipeline to be re-run whenever the model architecture or corpus changes.
% While effective in some cases, this paradigm suffers from fundamental drawbacks~(Sec.~\ref{sec:revisit}): Crucially, the 'value' of a data sample is not static; it changes as the model learns. Offline selection freezes this value based on a proxy model, creating a mismatch with the training model's evolving needs. Consequently, offline subsets often overfit to a specific benchmark, and once the architecture or corpus changes, previously curated data quickly becomes suboptimal, forcing practitioners to restart the pipeline. % In short, offline methods are brittle and introduce “hard cuts” that discard data diversity.

% In contrast, we propose a shift to \textit{online reweighting}, which dynamically adjusts sample contributions during training by weighting losses, without modifying the underlying dataset. This preserves data diversity compared to offline's “hard cuts”  intelligently controlling each sample's “step size” in parameter updates to adapt to evolving model states.

In contrast, we propose a paradigm shift to \textit{online reweighting}, which fundamentally differs from offline selection not just in \textit{when} the signal is computed, but also in \textit{how} the data is used. Instead of hard-filtering a subset based on static scores, online reweighting retains the full dataset to preserve diversity, but dynamically modulates each sample's contribution via loss weighting. This preserves data diversity compared to offline's ``hard cuts''  while intelligently controlling each sample's ``step size'' in parameter updates to adapt to evolving model states.

To solve the limitation of offline data curation, we propose a unified online reweighting framework, \textbf{ADAPT} (\textbf{A}daptive \textbf{D}ata reweighting for \textbf{P}retraining and Fine\textbf{T}uning), which aligns training samples with a validation set via similarity-based weighting. Our method incurs nearly zero additional overhead yet achieves superior cross-benchmark generalization. Under a unified evaluation protocol, it consistently outperforms both offline selection/mixing strategies and prior online reweighting methods, across large-scale pretraining and instruction tuning setups.

% To solve the issue of overfitting, we presented a framework that dynamically weights training samples based on their alignment with a validation set via different similarity calculations, ensuring that sample importance adapts continuously rather than being fixed. We evaluate on large-scale pretraining corpus and instruction data for post-training. 
% Building on this, we introduce a learnable threshold mechanism that adaptively regulates sparsity, enabling stable reweighting dynamics and improved out-of-domain generalization.
% Our analysis shows that with careful design, online reweighting can be realized with nearly zero additional overhead, while yielding superior cross-benchmark generalization [by X\%]. % is often believed to provide an efficient and generalizable recipe for LLM training. \todo{for pretraining, the case is the same. 

\paragraph{Our contributions.} 
This work makes the following contributions:

\begin{enumerate}[leftmargin=*]
    \item \faMedapps ~\textbf{Formalization and unification of dataset curation pipelines. } We present a unified view of data curation, showing that data selection, mixing, and balancing can all be expressed as special cases of \emph{data reweighting}. This perspective shifts the focus from offline pre-processing to dynamic, in-training adjustment of data weights, which offers a fair protocol to view and evaluate them through a unified lens of \emph{online data reweighting}. 
    % We design a framework that generalizes and unifies each of these stages as a form of reweighting. For example, we will treat filtering as assigning a weight of “zero” to low-quality data, and quality signals as upweighting certain documents. 
    % \textbf{A unified perspective.} We reformulate data selection, mixing, and balancing as instances of online data reweighting, shifting the focus from pre-training subset choice to dynamic in-training weight adjustment.
    % Based on this framework, we explore trade-offs % inherent in each stage, such as quality-aware deduplication where high-quality documents are allowed to be duplicated more than once. 
    \item \faPenFancy ~\textbf{Online sample-level reweighting method.} We introduce a new online global reweighting algorithm that is effective under this unified formalization and protocol. It employs an adaptive per-sample learning rate guided by sample quality signals, without explicitly altering the dataset size, while dynamically adjusting sample weights during training. % (ii) an adaptive per-sample learning rate with a \emph{learnable threshold} to regulate sparsity. Together, these mechanisms enable stable training and improved generalization across benchmarks. 
    \item \faRulerCombined ~\textbf{Better performance and generalization across tasks.} %We propose a new evaluation protocol that measures efficiency under \emph{total FLOPs} (including selection overhead) and tests generalization across both in-domain and out-of-domain benchmarks. This protocol reveals overlooked weaknesses of offline methods and highlights the strengths of online reweighting. % We design a stable update mechanism that controls the learning-rate range of weight adaptation and introduce a learnable threshold to regulate sparsity, preventing oscillations and collapse
    On both instruction tuning and large-scale pretraining experiments, our framework achieves up to 7.2\% higher accuracy per FLOP than offline baselines including LESS~\citep{xia2024less}, DoReMi~\citep{doremi}, and RegMix~\citep{liu2024regmix}, and other online reweighting methods~\citep{sow2025dynamic}, and demonstrates 6.1\% stronger out-of-domain generalization across benchmarks.
\end{enumerate}

% next:v5

\section{Background and Related Work} \label{sec:related_work}
% For example, pretraining dataset curation typically involves many stages that include deduplication, various ad-hoc filters to remove noisy and undesirable content, and heuristic or intuition-based reweighting of sources. Such curation is often applied to large-scale web scrapes. 

% mention online mixing
\subsection{Data curation for large language models}
% \paragraph{Data curation for large language models.}
Researchers and engineers usually rely on web crawls to gather large datasets for training large language models~\citep{gpt3, redpajama, penedo2024finewebdatasetsdecantingweb, tang2024txt360, bai2023qwen, kandpal2025common}. However, these crawls often include a substantial amount of low-quality or irrelevant content, which makes data curation necessary to build high-quality training sets.
Most data curation efforts focus on methods for improving model performance~\citep{raffel2020exploring, gpt3, gopher, penedo2023refinedweb, dolma}, including
filtering by language~\citep{ raffel2020exploring, xue2020mt5}, heuristic-based filtering~\citep{pile, gopher, penedo2023refinedweb, dolma}, quality filtering~\citep{du2022glam, xie2023data},
data deduplication~\citep{Lee2021DeduplicatingTD} and mixing~\citep{xie2023doremi, cerebras2023slimpajama, albalak2023efficient}.
% While prior work examines a limited set of filters, we conduct the largest public investigation of data curation, resulting in a strong \hqpool{} dataset. Since the initial release of DCLM, newer works such as WebOrganizer \citep{wettig2025organize}, Nemotron-CC \citep{su2024nemotroncc}, and Olmo-2 \citep{olmo20252olmo2furious} have also built upon our benchmark or curation strategies to further advance the state-of-the-art for LLM pre-training datasets.
However, current approaches to curating such datasets are generally ad-hoc. We aim to develop a principled and automated method for data curation that can also unify different processing stages. 
% 2. existing methods. forest of terminologies, mixing, balancing, selection, reweighting. motivate the need for unification; 

% 3. pretraining v.s. finetuning. they are separately studied in two branches but in this work we forge a common ground for them by proposing a method applicable for the two?

% Current practice for high-quality large scale data: FineWeb (no mixing), TxT360 (yes), RedPajama (no mixing), Common Pile (yes), DCLM, Dolma

% Pretraining data is not just about filtering, but involves a lot more stages. 

% data mixing

\paragraph{Pretraining data curation. }
Several recent studies \citep{xie2023doremi,chen2023skill,fan2023doge,thakkar2023self} have explored various reweighting techniques to enhance the generalization and efficiency of language models pretraining. For instance, \cite{xie2023doremi} and \cite{fan2023doge} are aiming at finding the optimal mixture of pretraining corpora to enhance performance across domains. \citet{chen2023skill} propose an ordered skill learning method for data selection measuring how effectively it teaches interdependent skills for continual pretraining and fine-tuning. Although effective, these works are aiming at the group level, whereas our work explores reweighting at the sample level, offering fine-grained control during model training dynamics. % Furthermore, we demonstrate that combining domain-level methods such as DoReMi \citep{xie2023doremi} or DoGE \citep{fan2023doge} with our instance-level reweighting methods results in improved performance across multiple domains.
Instance-level reweighting has been used in pretraining settings of LLMs \citep{chen2024take,jiang2024importance}, where each sample per mini-batch is weighted over how individual samples are treated based on their loss values. %\citet{jiang2024importance} boost the self-improvement abilities of LLMs by employing sample reweighting to filter out self-generated data that have correct answers but exhibit high distribution shifts. \cite{chen2024take} reweight individual samples during continual training/instruction-tuning to focus on medium-loss samples. 
In contrast, our work studies the effects of various adaptive learning rates with different quality signals considering the model states to enhance both performance and generalization for LLMs pretraining and fine-tuning regimes. 
\paragraph{Instruction data curation. }
% \paragraph{High-quality instruction tuning data for LLMs.}
%High-quality instruction tuning datasets can substantially enhance the large language models. 
Research has demonstrated that prioritizing data quality and diversity over quantity is more helpful for instruction-following capabilities~\citep{cao2023instruction,chen2023alpagasus,bukharin2023data,du2023mods, liu2023makes, li2023one}.
Instruction tuning data includes task-based datasets curated from traditional NLP tasks~\citep{wang2022super,sanh2022multitask,wei2022finetuned,longpre2023flan}, and open-ended  datasets~\citep{alpaca,DatabricksBlog2023DollyV2,kopf2023openassistant,xu2023wizardlm, mukherjee2023orca,zhou2023lima,ding2023enhancing}. In our work, we focus on adaptively adjusting the data weight during instruction tuning without explicitly reducing the number of instruction data samples to obtain a model with better generalization. 

% \paragraph{Commonly used quality signal for data curation.}
\subsection{Commonly used quality signals for data curation}
\label{sec:offline_curations}
%Such instruction data typically comprises two varieties: (1) datasets assembled from standard natural language processing challenges~\citep{wang2022super,sanh2022multitask,wei2022finetuned,longpre2023flan}, and (2) datasets for comprehensive instruction adherence that encompass diverse subject matters~\citep{alpaca,DatabricksBlog2023DollyV2,kopf2023openassistant,xu2023wizardlm, mukherjee2023orca,zhou2023lima,ding2023enhancing}.  Additionally, ~\citep{han2023context, lin2023unlocking} have illustrated that contextual learning by itself can markedly improve instruction compliance. Our contribution extends this discourse by curating superior, appropriate data from an extensive array of accessible datasets to foster specific model abilities.

%Increasing data quality and diversity instead of quantity has been shown to more effectively induce instruction following abilities~\citep{cao2023instruction,chen2023alpagasus,bukharin2023data,du2023mods, liu2023makes, li2023one}. Moreover, ~\citep{han2023context, lin2023unlocking} have demonstrated that in-context learning alone can significantly advance instruction following. Our work adds to this narrative by selecting high-quality, relevant data from a large pool of available datasets to induce a particular capability in the model.

\begin{itemize}[leftmargin=10pt]
    \item \textbf{Lexical Similarity (BM25)}~\citep{silva2024improving} quantifies term-based overlap between training and validation data through sparse retrieval scoring:
    \begin{equation}
    s_{\text{BM25}}(x) = \frac{1}{|\mathcal{D}_{\rm val}|}\sum_{v \in \mathcal{D}_{\rm val}} \text{BM25}(x, v).
    \end{equation}
    This metric captures surface-level textual similarity without semantic understanding.

    \item \textbf{Semantic Similarity (Embedding)}~\citep{rubin2021learning} measures dense representation alignment using pretrained encoders:
    \begin{equation}
    s_{\text{Embed}}(x) = \frac{1}{|\mathcal{D}_{\rm val}|}\sum_{v \in \mathcal{D}_{\rm val}} \cos\big(\phi(x), \phi(v)\big),
    \end{equation}
    where $\phi(\cdot)$ denotes a frozen embedding model. This approach captures semantic proximity beyond lexical overlap.

    \item \textbf{Distributional Alignment (Perplexity)}~\citep{antonello2020selecting} evaluates likelihood under a reference language model $\theta_0$:
    \begin{equation}
    s_{\text{PPL}}(x) = -\log P_{\theta_0}(x).
    \end{equation}
    Lower perplexity indicates stronger distributional alignment with the reference corpus.

    \item \textbf{Gradient-Based Influence}~\citep{xia2024less} estimates training utility through first-order approximation of validation loss reduction:
    \begin{equation}
    s_{\text{Grad}}(x) = \big\langle \nabla_\theta \ell(x; \theta_0), \nabla_\theta \ell(\mathcal{D}_{\rm val}; \theta_0) \big\rangle,
    \end{equation}
    where $\theta_0$ denotes a proxy model. This metric directly quantifies how training on $x$ influences performance on the validation set $\mathcal{D}_{\rm val}$.
\end{itemize}
\section{Unified Formalization of Data Curation} 
\label{sec:unification_formalization}
\paragraph{Setup.}

Given two datasets: the {\em train} set $\mathcal{D}_{\rm train}$ and the {\em validation} set $\mathcal{D}_{\rm val}$. Usually, the size of the validation set is much smaller than the train set. When $\mathcal{D}_{\rm val}$ is sampled from the distribution of downstream test data, the validation dataset $\mathcal{D}_{\rm val}$ is considered as {\em in-domain}. The train dataset $\mathcal{D}_{\rm train}$ on the other hand consists of both in-domain and {\em out-of-domain} samples. This is the case when web crawling is used to collect training data from the whole internet. 
Our goal is to subsample or weight the train set $\mathcal{D}_{\rm train}$ under the guidance of the validation set $\mathcal{D}_{\rm val}$, so that the model training is less affected by the out-of-domain samples.

\subsection{Design Space of Data Quality Signal} 
% put this as the first subsection of this section, giving a smooth transition from the related work. 
% Existing work evaluates several scoring functions that estimate the quality of training data relative to a validation set or a reference model. Here we only consider ML-based heuristic method those without using proxy models, of which complexity is dependent on specific training recipes. 
% Within our unified framework, these scoring functions define $s(u)=f(u;\mathcal{D}_{\rm val})$,  which can be applied at either the example level ($u=x$) or the domain level ($u=D_i$). 
% The resulting scores can then be used for different curation paradigms: offline \emph{selection} (binary weights), offline \emph{mixing} (domain-level proportions), or online \emph{reweighting} (continuous per-example weights). 

% Here use the itemize maybe better, like this:
%We present a unified framework for data quality assessment where scoring functions take the form $s(x) \equiv s(x; \theta, \mathcal{D}_{\rm val})$, parameterized by a reference model $\theta$ and validation set $\mathcal{D}_{\rm val}$. 
%The input $u$ can represent either individual examples ($u = x$) or entire domains ($u = D_i$), enabling application across all curation paradigms.
In Sec.~\ref{sec:related_work}, we present a unified framework for data quality assessment where \emph{scoring function} takes the form $s(x) \equiv s(x; \theta, \mathcal{D}_{\rm val})$ that assigns a quality signal for each data example $x$, which optionally depends on a (proxy) model $\theta$ and the validation set $\mathcal{D}_{\rm val}$. %In the following, we list a few representative implementations of $s(x)$.

\subsection{Offline Data Curation}

% \todo{assumption}. 
The scores are employed once before training (e.g., with a proxy model $\theta_0$ and fixed validation set $\mathcal{D}_{\rm val}$). 
The resulting weights $\{w(x)\}_{x \in \mathcal{D}_{\rm train}}$ are fixed and used to make a decision to pass which training examples to the real training stage.

\begin{itemize}[leftmargin=10pt]
 \item \textbf{Data Selection (sample-level binary weights)}
Data selection (or filtering) removes part of the training corpus before pretraining begins. 
It keeps data examples with quality signal above a certain threshold $\tau$:
\[
\mathcal{D}^{\prime}_{\rm train} = \{x \in \mathcal{D}_{\rm train} \mid s(x) \geq  \tau \}.
\]
%With carefully chosen $s(x)$ and $\tau$, the new training set $\mathcal{D}^{\prime}_{\rm train}$ is considered of high quality, leading to a better model.
From a data reweighting perspective, filtering data points is equivalent to assigning binary weights of the form $w(x) = \mathbf{1}[ s(x) \geq \tau ] \in \{0, 1\}$.

% \paragraph{Data Mixing (domain-level fractional(interger?) weights)} 
 \item \textbf{Data Mixing (domain-level fractional weights)} 
 \label{sec:data_mix}
Data mixing can be seen as a coarse-grained data reweighting method that operates at the domain level. 
In other words, data points within the same domain receive the same quality score. Pulling all domain scores $\{s_d\}_{d \in {\rm domains}}$ and normalize them to obtain data mixing probability for each domain $d$
\[
w_d = \frac{g(s_d)}{\sum_{d'} g(s_{d'})},
\]
where $g$ transform the score to be non-negative with a common choice being $g \colon s \mapsto \exp(s)$. As such, $w_d$ determine how much of a total training budget $B$ is allocated to each domain: $B_d = w_d B$. Alternatively, training can be implemented by sampling domains with probability $w_d$ (probability mixing) or by assigning a fixed quota $B_d$ (quota mixing).  
Since $w_d$ is decided before training, mixing is an \emph{offline} operation that alters the effective number of examples each domain contributes.
\end{itemize}

\subsection{Online Data Curation}
% The scores depend on the evolving model state $\theta_t$ and possibly dynamic validation set. 
% The weights $\{w_t(x)\}_{x \in \mathcal{D}_{\rm train}}$ might be time dependent, modulating \emph{gradient contributions} without reducing the volume of the training set $\mathcal{D}_{\rm train}$.
% Paul:
In this setting, the scoring function depends on the evolving model state $\theta_t$ and, in some cases, on a dynamic validation set.  
The resulting weights $\{w_t(x)\}_{x \in \mathcal{D}_{\rm train}}$ evolve throughout training, adjusting the \emph{gradient contributions} of examples while preserving the full volume of the training set $\mathcal{D}_{\rm train}$.

\paragraph{Data Reweighting (sample-level fractional weights)}
Data reweighting assigns normalized weights to examples while keeping the training set size unchanged.   
Similar to Sec.\,\ref{sec:data_mix}, the quality score for example $x$ is transformed to a non-negative weight: $w(x) = g(s(x))$, 
which scales their contribution to the loss:
\[
\mathcal{L}^*(\theta) = \frac{1}{Z} \sum_{x \in \mathcal{D}} w(x)\, \mathcal{L}(\theta; x),
\qquad Z = \sum_{x \in \mathcal{D}} w(x).
\]
Equivalently, from a stochastic gradient descent perspective, the stochastic gradient w.r.t. $x$ is scaled by $\frac{w(x)}{Z}$, which plays a role of \emph{per-sample} learning rate in addition to the \emph{global} learning rate $\eta$:
\[
\theta_{t+1} = \theta_t - \eta \frac{w(x)}{Z}\,\nabla_\theta \ell(\theta_t; x).
\]
Since the total number of training examples remains unchanged, only their relative contribution to parameter updates is modified, data reweighting is inherently an \emph{online} method.

% \paragraph{Offline vs.\ Online distinction.}
% The difference now becomes clear in terms of the usage of data weights:
%\begin{itemize}
%    \item \textbf{Offline methods} (selection, mixing): 
\vspace{-0.3cm}
\subsection{Unified Evaluation of Total FLOPs (Offline vs.\ Online)}
The unification enables us to compare different methods under the same FLOPS calculation framework.
We evaluate curation methods under a \textit{cost-aware} metric that accounts for all computation spent to obtain and use the curated data. Let $B$ denote a training budget (tokens or update steps).
\vspace{-0.3cm}
\paragraph{Total FLOPs for offline data curation}
Offline curation modifies the \emph{amount of data that participates in training} before training starts.
Its total compute is
\begin{equation}
\label{eq:offline-total}
\mathsf{F}^{\mathrm{off}}_{\mathrm{total}}
\;=\;
\underbrace{\mathsf{F}^{\mathrm{off}}_{\mathrm{prep}}(\mathcal{D}; f)}_{\text{data scoring as preprocessing}}
\;+\;
\underbrace{\mathsf{F}_{\mathrm{train}}\!\big(\mathcal{D}',\,B'\big)}_{\text{model training on the sampled subset}},
\end{equation}
where (i) $\mathsf{F}^{\mathrm{off}}_{\mathrm{prep}}$ includes any corpus-wide scoring, proxy-model passes, retrieval or filtering necessary to construct $\mathcal{D}'$; 
(ii) $\mathcal{D}'\subseteq\mathcal{D}$ is the retained subset (for selection), or an \emph{effective} subset induced for data mixing; 
(iii) $B'$ is the effective training budget after curation. %(e.g., $B'=\sum_{i} p_i B$ under quota mixing, or the expected number of examples seen from $\mathcal{D}'$ under probability mixing).
By construction, offline selection uses \emph{binary} example weights (keep/drop), and offline domain mixing changes per-domain sampling rates, thereby reducing or reallocating the volume of data seen during training.
\vspace{-0.3 cm}
\paragraph{Total FLOPs for online data curation}
Online curation \emph{does not change the amount of training data}; instead, it modulates each example's contribution \emph{during training}. Its total compute is
\begin{equation}
\label{eq:online-total}
\mathsf{F}^{\mathrm{on}}_{\mathrm{total}}
\;=\;
\underbrace{\mathsf{F}_{\mathrm{train}}\!\big(\mathcal{D},\,B\big)}_{\text{model training on the full corpus}}
\;+\;
\underbrace{\mathsf{F}^{\mathrm{on}}_{\mathrm{metrics}}}_{\text{on-the-fly features}},
\end{equation}
where $\mathsf{F}^{\mathrm{on}}_{\mathrm{metrics}}$ denotes lightweight, in-training computations needed to obtain scores (e.g., using current logits, per-example loss, or gradient norms).
Because scoring is amortized inside the training loop and no data are removed, $\mathsf{F}^{\mathrm{on}}_{\mathrm{metrics}}$ is typically modest relative to $\mathsf{F}_{\mathrm{train}}$.

% \paragraph{Unifying the three paradigms under cost.}
% Given these definitions:
% \begin{itemize}
%  \item \textbf{Data selection (offline)} uses binary weights $w(x)\!\in\!\{0,1\}$ to form $\mathcal{D}'$, decreases the training volume, and incurs $\mathsf{F}^{\mathrm{offline}}_{\mathrm{prep}}$ for filtering/scoring; its efficiency must be judged by Eq.~\eqref{eq:offline-total}, not by training FLOPs alone.
% \item \textbf{Data mixing (offline, domain reweighting)} sets domain proportions $\mathbf{p}$ that directly control how many tokens/steps each domain contributes, thus changing the effective data volume per domain; its efficiency is also evaluated by Eq.~\eqref{eq:offline-total}.
%  \item \textbf{Data reweighting (online)} keeps the \emph{full} training volume ($\mathcal{D}, B$ unchanged) and scales example contributions in the loss/updates; its efficiency is given by Eq.~\eqref{eq:online-total}.
% \end{itemize}

%\paragraph{Practical implication.}
% Under equal reported “training FLOPs”, offline methods can appear deceptively efficient if the preprocessing FLOPs $\mathsf{F}^{\mathrm{offline}}_{\mathrm{prep}}$ (e.g., proxy model passes or large-scale scoring) are omitted. 
Conversely, online reweighting maintains the full training signal while adding only minimal in-loop metrics computation, providing a clearer accuracy--compute trade-off when measured by \emph{total} FLOPs.
\vspace{-1.5em}
\section{Revisiting Offline Data Selection}
\label{sec:revisit}
%\vspace{-0.5cm}
In this section, based on the evaluation protocol derived from the unification above, we revisit the accuracy--FLOPs trade-off of existing offline data selection methods. % \todo{add reference to flops estimation} \todo{add reference to baseline descriptions in sectioin 2. } \todo{experimental setup}

% Despite the growing interest in data selection for pretraining large language models, most existing approaches still rely on \emph{offline paradigm}, where a fixed subset of training data is chosen with respect to a given validation benchmark before model training begins. Although this paradigm has delivered strong empirical results, we argue that it suffers from several fundamental \textbf{pitfalls}. Our unified evaluation protocol, which measures efficiency under total FLOPs and considers both in-domain and out-of-domain performance, allows us to expose these weaknesses that were largely hidden under previous evaluation practices.
\vspace{-0.21cm}
\begin{figure}[htbp]
  \centering
  \floatsetup{justification=centering}
  \ffigbox[\FBwidth]{%
    \begin{subfloatrow}[2]
      \ffigbox{\includegraphics[width=1.0\linewidth]{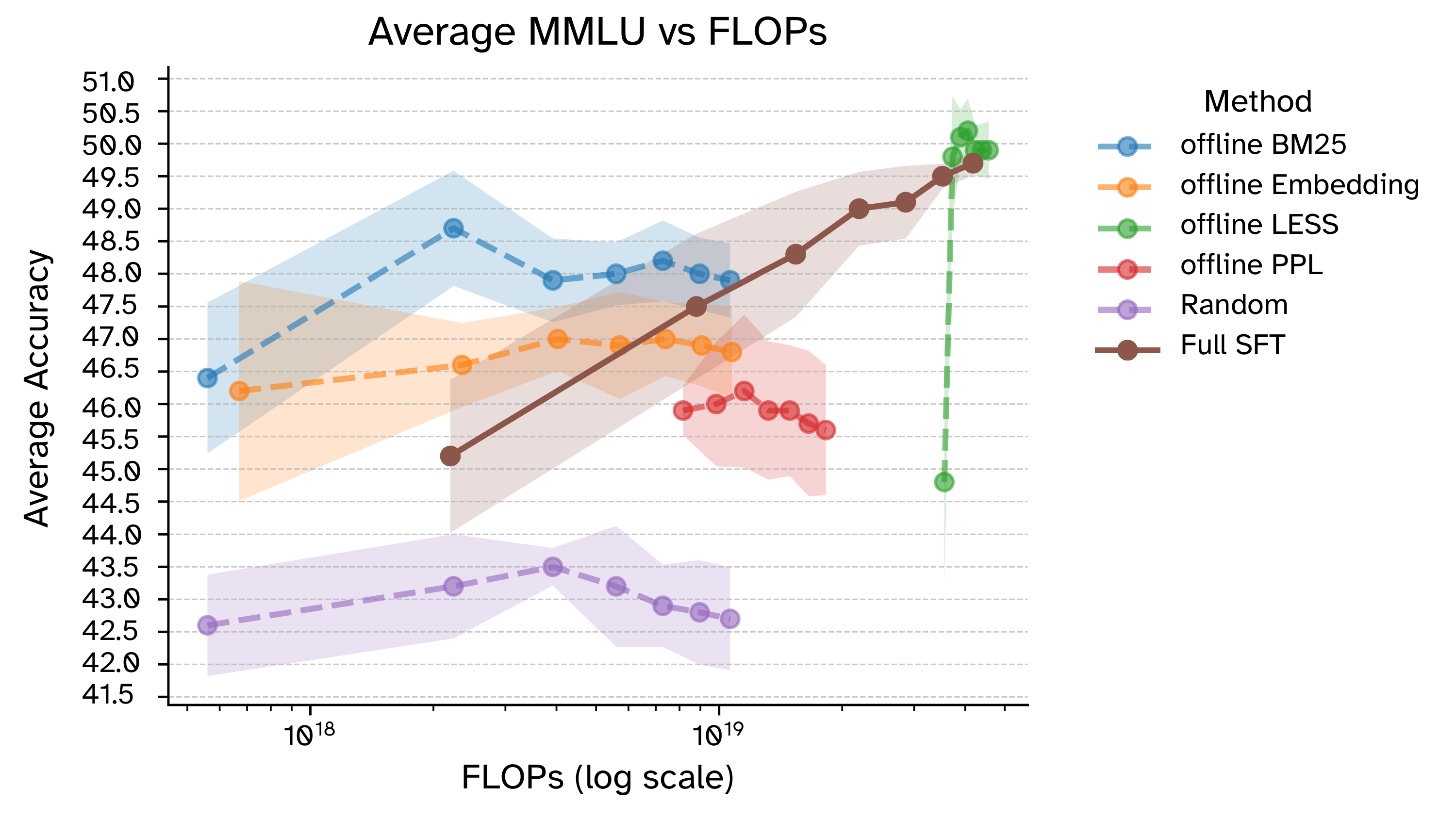}}%
      {\caption{In-domain evaluation: MMLU validation set $\rightarrow$ MMLU benchmark.}
      \label{fig:observation_mmlu}}
      \ffigbox{\includegraphics[width=1.0\linewidth]{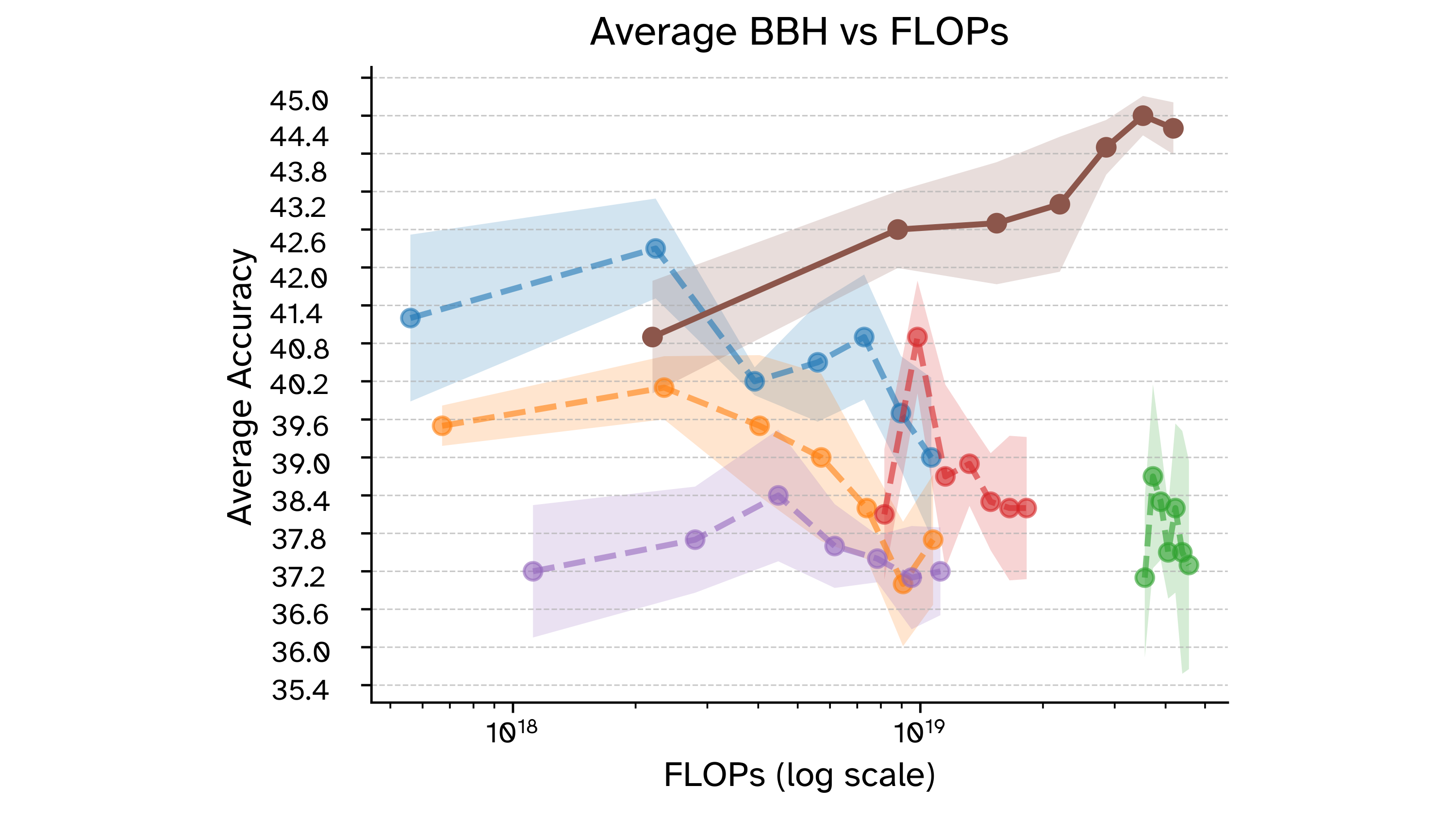}}%
      {\caption{Out-of-domain evaluation: MMLU validation set $\rightarrow$ BBH benchmark.}
      \label{fig:observation_bbh}}
    \end{subfloatrow}
    \vspace{-0.8em}
  }{%
    \caption{Efficiency--accuracy tradeoff of different data selection methods and full finetune method under our proposed \textbf{total FLOPs} metric. Selection methods (e.g., \texttt{LESS}) appear competitive under in-domain evaluation, but their advantage vanishes or even reverses in out-of-domain settings. In contrast, (\texttt{full sft}) remains consistently strong across domains.}
    \label{fig:observation}
  }
\end{figure}
\vspace{-0.523cm}

\paragraph{Observation.} Figure~\ref{fig:observation} illustrates the limitation of offline data selection methods:~\textbf{Overfitting to the  validation benchmark.} Models trained on offline-selected data with MMLU~\citep{mmlu} as validation set perform well on the MMLU benchmark (Figure~\ref{fig:observation_mmlu}), but the same models generalize poorly to the BBH benchmark~\citep{bbh} (Figure~\ref{fig:observation_bbh}). This issue is especially severe when using LESS~\citep{xia2024less}. This reveals that offline methods often \emph{overfit to the chosen validation task}, lacking a true generalization. % [be concrete: how much in domain, how much out of domain, drop x\%] 
However, we observed that 
vanilla full dataset training demonstrates more stable performance in both benchmarks. One possible reason is that offline data curation, in which we directly change the number of data involved in the training by repetition, would cause the model to replace generalization ability with memorization~\citep{hernandez2022scaling}. 
% generalization performance between benchmark compared to other offline data selection baselines and demonstrates comparable generalization with vanilla full dataset instruction tuning.

	% \item ~\textbf{Hidden computational cost.}  
    % \item \textbf{Brittleness across model generations.}  
    % Data subsets selected offline are tied to a specific model state. Once the architecture or pretraining corpus changes, the previously chosen subset becomes suboptimal, forcing practitioners to re-run the entire pipeline from scratch.
    %
    % However, prior evaluations only reported training FLOPs, ignoring the cost of proxy models and large-scale filtering pipelines. Under our \emph{total FLOPs} metric, methods such as \texttt{LESS} fall strictly below [X\%] simple baselines like \texttt{BM25}, exposing their inefficiency once the full cost is accounted for.
    % \item ~\textbf{Misconception about online methods.}  

% Together, these pitfalls highlight that the current offline paradigm is neither cost-effective nor generalizable. By introducing a unified evaluation protocol that accounts for both total FLOPs and out-of-domain performance, we uncover the hidden strength of online data reweighting methods, which were largely ignored under previous evaluation practices.

In summary, the offline paradigm has severe limitations, being neither cost-effective nor generalizable. This motivates us to explore an online alternative, which may alleviate memorization while retaining more data to enhance generalization. In the next section, we leverage these insights to design an \emph{online reweighting method}.
 % next: v4
\section{Adaptive Learning Rate for Online Data Reweighting}
\vspace{-.5em}
% method outline
% online sample-level reweighting: sim (train, val)
% BM25 (BOW): fixed learning rate, fixed threshold $f(\theta_0, threshold=constant)$ 
% RDS (embedding): adaptive learning rate, fixed threshold f(theta_t, threshold=constant)
% STREAM (gradient): adaptive learning rate, adaptive threshold f(theta_t, threshold)
% Shell method. meta learning per-sample 

% shorten the introduction paragraph
The limitations of offline data selection suggest that a different paradigm is needed: one that is generalizable, and adaptive to the evolving state of the model. To this end, we propose an \textbf{online data reweighting framework} that is also suitable for data selection, data mixing with a unified formalization. 
Unlike offline methods that commit to a fixed subset before training begins, our framework dynamically adjusts data weights as the model learns, so it naturally adapts to new model states without re-running the data preprocessing and training pipeline from scratch. % (ii) it eliminates the hidden overhead of proxy models by integrating reweighting into training, and (iii) it maintains strong generalization across benchmarks by preventing overfitting to a single validation set. 
%
% In the following subsections, we first present the unified formulation of data reweighting, then describe its efficient implementation, and finally discuss how it subsumes prior approaches such as data selection and mixing as special cases.

\subsection{Per-Sample Learning Rate Update}

We cast data selection and mixing as \emph{online data reweighting}. 
At training step $t$, given a minibatch $B_t \subset \mathcal{D}$, we update model parameters as
\begin{equation}
    \theta_{t+1} = \theta_t - \eta \sum_{i \in B_t} w_t(i)\,\nabla_\theta \ell\big(f_\theta(x_i), y_i\big),
\end{equation}
where $w_t(i) \geq 0$ denotes a dynamic weight assigned to sample $i$. 
Here, $w_t(i)$ directly scales the gradient contribution of sample $i$, and thus acts as a 
\emph{per-sample learning rate multiplier}: larger weights amplify the effective step size on informative examples, 
while smaller weights downweight less useful ones. 

In our framework, weights are derived from scoring functions $s_t(v,i)$ that compare a training example $i$ with validation/query points $v \in \mathcal{D}_{\rm val}$. 
These scores are aggregated across $v$, normalized within the current batch or pool, and transformed by a smooth gating function to produce the final weights $\widehat{w}_t(i)$. 
To ensure stability, weights are clipped to prevent excessively large effective learning rates and avoid gradient explosion. 
We now describe our score functions. % \todo{instead of explicitly using the name, we should define it from scratch and cite references.}
% \paragraph{BM25 scoring.}.
% \vspace{-.4cm}
\subsection{Model-Agnostic Per-Sample Learning Rate Update}
% \begin{itemize}[leftmargin=10pt]
\textbf{\ours{}-BM25} quantifies term-based overlap between training and validation data through sparse retrieval scoring:
    \begin{equation}
    s_{\text{BM25}}(x) = \frac{1}{|\mathcal{D}_{\rm val}|}\sum_{v \in \mathcal{D}_{\rm val}} \text{BM25}(x, v).
    \end{equation}
    This metric captures surface-level textual similarity without semantic understanding.
    We use the standard BM25 score $s_{\mathrm{BM25}}(v,i)$ over sparse token matches between query $v$ and example $i$. Aggregated scores are normalized to obtain target weights.
% \vspace{-.4cm}
\subsection{Per-Sample Learning Rate Update with Model States}

\textbf{\ours{}} measures alignment using the model's own dense representations rather than a frozen encoder~\citep{ivison2025large}. 
For an input $x$ with last-layer hidden states $\{h_i\}_{i=1}^L$, we compute a position-weighted mean pooling:
\begin{equation}
w_i = \frac{i}{\sum_{j=1}^L j}, 
\qquad 
\phi(x) = \sum_{i=1}^L w_i h_i,
\end{equation}
where later tokens receive higher weights to counteract the causal mask bias of decoder-only models.  
We then define the similarity score:
\begin{equation}
s_{\text{\ours{}}}(x) = \frac{1}{|\mathcal{D}_{\rm val}|}\sum_{v \in \mathcal{D}_{\rm val}} \cos\big(\phi(x), \phi(v)\big).
\end{equation}
Equivalently, we can instantiate a representation-based scorer 
$s_{\mathrm{\ours{}}}(v,i)=\cos\big(\phi(v),\phi(i)\big)$, 
where $\phi(\cdot)$ is the weighted hidden representation. 

We use a sigmoid function with temperature scaling to produce \textit{absolute} weights that are independent of the batch composition:
\begin{equation}
w_t(i) = \sigma\left(\frac{s_{ADAPT}(x_i)}{\max(\tau, \epsilon)}\right) = \frac{1}{1 + \exp(-s_{ADAPT}(x_i) / \max(\tau, \epsilon))}
\end{equation}

where $\sigma(\cdot)$ denotes the sigmoid function, $\tau > 0$ is a temperature hyperparameter (default $\tau = 1.0$), and $\epsilon > 0$ is a small constant (e.g., $\epsilon = 10^{-8}$) used for numerical stability. The temperature parameter $\tau$ controls the steepness of the sigmoid: larger values produce flatter weight distributions, while smaller values create sharper distinctions between high and low similarity samples.

\textbf{Global Weighting via Sigmoid:} The sigmoid transformation maps similarity scores to absolute weights in the interval $[0, 1]$ without requiring normalization across samples in the batch. This design choice ensures that the weight assigned to each sample $x_i$ depends solely on its similarity score $s_{ADAPT}(x_i)$ relative to the anchor set, rather than its rank within the current mini-batch. Consequently, a sample with a given similarity score receives the same weight regardless of whether it appears in a high-quality or low-quality batch, making the weighting mechanism robust to batch-level variations in data quality. This contrasts with normalized weighting schemes (e.g., softmax normalization) where weights are relative to other samples in the same batch, potentially amplifying or dampening effects based on batch composition.

\textbf{Embedding Normalization:} We apply L2 normalization to embeddings $\phi(x)$ and $\phi(v)$ before computing cosine similarities to ensure scale-invariant similarity measurements. Specifically, for each embedding vector, we compute $\phi(x) \leftarrow \phi(x) / \max(\|\phi(x)\|_2, \epsilon)$ where $\epsilon$ prevents division by zero. This normalization step is distinct from weight normalization and serves to standardize the embedding space for similarity computation.

\textbf{Online Validation Embedding Updates:} To maintain alignment with the evolving model representations, anchor embeddings $\{\phi(v)\}_{v \in \mathcal{D}_{val}}$ are refreshed every $R$ training steps via forward passes on the current model parameters $\theta_t$. This online update mechanism ensures that similarity scores reflect the model's current representation space rather than outdated embeddings from earlier training stages. The refresh interval $R$ is a hyperparameter that balances computational efficiency with representation freshness. The complete algorithm is summarized in Algorithm~\ref{algo:rds-online}.

While gradient-based methods (e.g., LESS) directly measure influence, gradients significantly fluctuate during early training and are computationally expensive to compute frequently. In contrast, semantic embeddings provide a smoother, more stable signal of relevance that evolves consistently with the model's representation space, offering a robust guide for reweighting without the noise of instantaneous gradient estimates.

% no meta
\section{Evaluation}
\subsection{Experimental Design}
\label{sec:exp_design}
\vspace{-.5em}
% Here we present a series of dataset ablations with the aim of developing a better understanding of how the quality signals introduced in Section 4.1.2 influence the downstream performance of language models trained on data filtered with different heuristics. More specifically, here we ask how do different quality filtering rules affect downstream performance? We strive for a broad evaluation and measure the performance on diverse downstream benchmarks and the language modeling objective on multiple domains.

%Our experiments address the following research questions:

%\begin{itemize}[noitemsep,topsep=0pt,parsep=2pt,partopsep=0pt]
 %   \item[\textbf{RQ1}] How does online data reweighting with \emph{heuristic per-sample learning rate update} perform compared to offline method? 
  %  \item[\textbf{RQ2}] How does online data reweighting with \emph{adaptive per-sample learning rate with learnable threshold} perform compared to offline method? 
  %  \item[\textbf{RQ3}] How does online data reweighting related to offline data mixing and selection? 
  %  \item[\textbf{RQ4}] How does online data reweighting perform in the large-scale pretrianing setup? % compared to offline method and other online baselin 
% \end{itemize}

\paragraph{Instruction Tuning Models and Data.}
\label{subsec:setup}
% \paragraph{Models and Data.} 
We use LoRA~\citep{hu2021lora} to fine-tune the base model \llamasmall{}~\citep{touvron2023llama2}.
Following the experimental setup in \cite{wang2023far}, we use the instruction tuning datasets including \flan~\citep{longpre2023flan},  \cott~\citep{wei2022chain}, \dolly~\citep{DatabricksBlog2023DollyV2} and \oasst~\citep{kopf2023openassistant}. The datasets do not contain any obvious in-domain data for the target queries.  % The datasets, comprising approximately 270K data points, vary widely in their format and underlying reasoning tasks, and % Table~\ref{tab:train_data} contains more details about the dataset, and more details can be found in \ref{app:train_data}.
We evaluate our method on \mmlu~\citep{hendrycks2020measuring}, multiple-choice dataset spanning 57 tasks and \bbh~\citep{suzgun2023challenging} from BIG-Bench selected to evaluate reasoning capabilities. For MMLU, we report 5-shot accuracy; for BBH, we report 3-shot exact match score. %, and \tydiqa~\citep{tydiqa}, multilingual question-answer dataset covering 9 typologically diverse languages. For MMLU, we report 5-shot accuracy; for BBH, we report 3-shot exact match score; and for TydiQA, we report 0-shot accuracy. % ~\mmlu{} consists of including elementary mathematics, US history, computer science, law, and more.
%
% ~\bbh{} is a collection of 27 challenging tasks from BIG-Bench selected to evaluate reasoning capabilities.
%
% ~\tydiqa{} is a multilingual question-answer dataset containing 9 typologically diverse languages. Given a question and a related passage, the task requires extracting the answer from the passage.
% Table~\ref{tab:eval_dataset} contains more details about these tasks. 
% Each dataset includes multiple subtasks, and each subtask comes with few-shot examples. 
% These examples are used as $\dval$ for data selection (\textsection\ref{sec:alg}) and as few-shot in-context learning demonstrations in evaluation. 
Appendix \ref{app:lora} contains more details on the training hyperparameter, Appendix \ref{app:train_data} and \ref{app:eval_data} contain more dataset details.
\vspace{-0.17cm}
\paragraph{Instruction Tuning Baselines.} We introduce instruction finetuning baselines in Sec.~\ref{sec:offline_curations}.

\paragraph{Pretraining Models and Data.}
% \subsection{Setup}\label{sec:setup}
We adopt \textbf{Tinyllama} architecture~\citep{tinyllama} with 120M parameters, with FlashAttention~\citep{flash} and Lit-GPT~\citep{litgpt}. More details about the training settings can be found at Appendix \ref{app:pretrain_details}.
Following prior works~\citep{llama,tinyllama,qrating,doremi}, we employ \textbf{SlimPajama}~\citep{llama,redpajama} as the text corpus, which is specifically curated for pre-training LLMs. 
All selections are performed on about 590M training files of SlimPajama, processed with Llama tokenizer~\citep{llama}. 
%
% Notably, QuRating~\citep{qrating} provides judgments on various properties of text samples in SlimPajama using a judge model trained based on GPT-3.5-turbo, which can be directly utilized in our implement.
%
% To capture the diversity dilemma and evaluate generic performance of pre-trained LLMs, we use seven commonsense reasoning tasks from the popular framework \textbf{Harness}~\citep{harness}, including four reading comprehension tasks~(\textbf{ARC-e}, \textbf{ARC-c}~\citep{allenai:arc}, \textbf{OBQA}~\citep{OpenBookQA2018}, and \textbf{BoolQ}~\citep{clark2019boolq}), and three physical world knowledge tasks~(\textbf{PIQA}~\citep{bisk2020piqa}, \textbf{HellaSwag}~\citep{zellers2019hellaswag}, and \textbf{WinoGrande}~\citep{sakaguchi2021winogrande}).
%
% Besides, we also employ another two tasks \textbf{MMLU}~\citep{MMLU}, and \textbf{BBH}~\citep{bbh} from Harness to evaluate the problem-solving capabilities for further clarify of our effectiveness.
%
We evaluate our method on a diverse set of 15 downstream benchmarks, following common practice in prior work such as \regmix\citep{liu2024regmix}.  
These tasks span a wide range of realistic settings, including: ARC-E \citep{arc_easy18}, ARC-C \citep{arc_easy18}, COPA \citep{copa20}, HellaSwag \cite{zellers2019hellaswag}, Lambada-S \citep{paperno2016lambada}, Lambada-O \citep{radford2019language}, LogiQA \citep{liu2020logiqa}, MultiRC \citep{multirc18}, OpenBookQA \citep{openbook18}, PiQA \citep{bisk2020piqa}, QQP \citep{qqp18}, RACE \citep{lai2017race}, SciQ \citep{sciq17}, Social IQA \citep{sap2019socialiqa}, WinoGrande \citep{sakaguchi2021winogrande}. The reported accuracy in \cref{tab:tinyllama_120M_50B} is measured in the 0-shot setting scored using the lm-eval-harness evaluation framework \citep{eval-harness}. 
%
% These benchmarks cover a diverse range of tasks, enabling a comprehensive evaluation of the real-world impact of \ours{}. %For each benchmark, we use normalized accuracy as the evaluation metric if provided by lm-eval-harness else we use regular accuracy. Notably, \ours also shows competitive performance across all downstream tasks even compared with \regmix, a task-aware method.
%
% See Section~\ref{sec:related} for detailed introduction of the dataset and tasks.
% \textbf{Model architecture.}~~~~
%
% All experiments and selection are implemented by PyTorch~\citep{pytorch} on platforms with 8 GPUs and 64 CPUs. 
%
% Except for Tinyllama, we also adopt OPT~\citep{opt} and Pythia~\citep{pythia} to verify the scalability of our method on model architecture as ablation study shown in Section~\ref{sec:ablation}.
%
% For feature extraction, we utilize the Contriever~\citep{contriever} with approximately 110M parameters to calculate feature representations of the text samples as defined in \eqref{eq:feature}. 
%
% We also experiment with other pre-trained models as feature extractors, including the text encoder of CLIP~\citep{clip} and GPT-2~\citep{gpt2}, as discussed in the ablation study of Section~\ref{sec:ablation}.
%
% See Appendix~\ref{app:model} for detailed introduction of model structures, as well as their training times.
\vspace{-6pt}

\paragraph{Pretraining Baselines.}
We compare \ours{} with \textbf{Uniform} selection and existing file selection methods for LLM pre-training, including \textbf{Doremi}~\citep{doremi} and \textbf{RegMix} \citep{liu2024regmix}. \ours{} requires an anchor set typically consists of examples in the evaluation distribution. To construct this set, we sample 50 validation examples from each of eight evaluation benchmarks: ARC-C, COPA, Lambada, MultiRC, PiQA, RACE, SciQ, and Social IQA.
%tbf{DSIR}~\citep{dsir}, QuRating~\citep{qrating}, \textbf{INGENIOUS}~\citep{ingenious}, and \textbf{D4}~\citep{d4}. 
%修改
%Since DSIR improves on Heuristic classification~\citep{gpt3,palm}, we present only DSIR results based on Wikipedia and Books. 
%
%For QuRating, based on the top judgment values, we select text samples for writing style (denoted as \textbf{QuRating-W}) and uniformly select across all styles (denoted as \textbf{QuRating-A}). 
%
%{For INGENIOUS, we utilize Contriever to extract features, which is more efficient than warmed-up model, as analyzed in Appendix~\ref{app:INGENIOUS}.}
%
% Additionally, for a comprehensive comparison, we include \textbf{Doremi}~\citep{doremi}, a recently proposed method that produces weights for pre-training on multiple text domains. 
%
For Doremi and Regmix we use the domain weights in \citep{lu2023chameleon} as the selection ratio of text samples in different domains in our experiment. For Uniform we use the same ratio to sample from each domain. We also compare \textbf{LinUpper}~\citep{sow2025dynamic}: an online sample reweighting strategy where the sample weight is proportional to the normalized loss but is capped at a predefined $\alpha$ value, ensuring that outliers do not dominate the training process. 
% See Appendix~\ref{app:baseline} for detailed implementation of the baselines. 
Due to the large cost of pretraining, for each method we sample 9B unique tokens from SlimPajama, and train for a total budget of 50B tokens, i.e., train for approximately 5.6 epochs.
%
% Notably, in our ablation studies, we also compare our method with \textbf{Full Data} pre-training, which refers to pre-training the LLMs on all training files in SlimPajama until the specified training budget is reached.
%

\vspace{9pt}
% \subsection{Main Results}
% \section{Evaluation}
\vspace{-0.2cm}
\subsection{Results for Instruction Tuning}
% \subsection{Online data reweighting with \emph{heuristic per-sample learning rate update} outperform offline methods~(\textbf{RQ1})}\label{sec:eval:momentum_noise_mutual_information}

%\begin{wrapfigure}{r}{0.5\linewidth}
%    \centering
%    \vspace{-1em}
%    \includegraphics[width=\linewidth]{figures/rds.png}
%    \vspace{-1em}
%    \caption{\small{Illustration of a model deployment plan.}}
%    \vspace{-1em}
%    \label{fig:rds}
%\end{wrapfigure}
\vspace{-0.4cm}
\begin{figure}[htbp]
  \centering
  \floatsetup{justification=centering}
  \ffigbox[\FBwidth]{%
    \begin{subfloatrow}[2]
      \ffigbox{\includegraphics[width=1.1\linewidth]{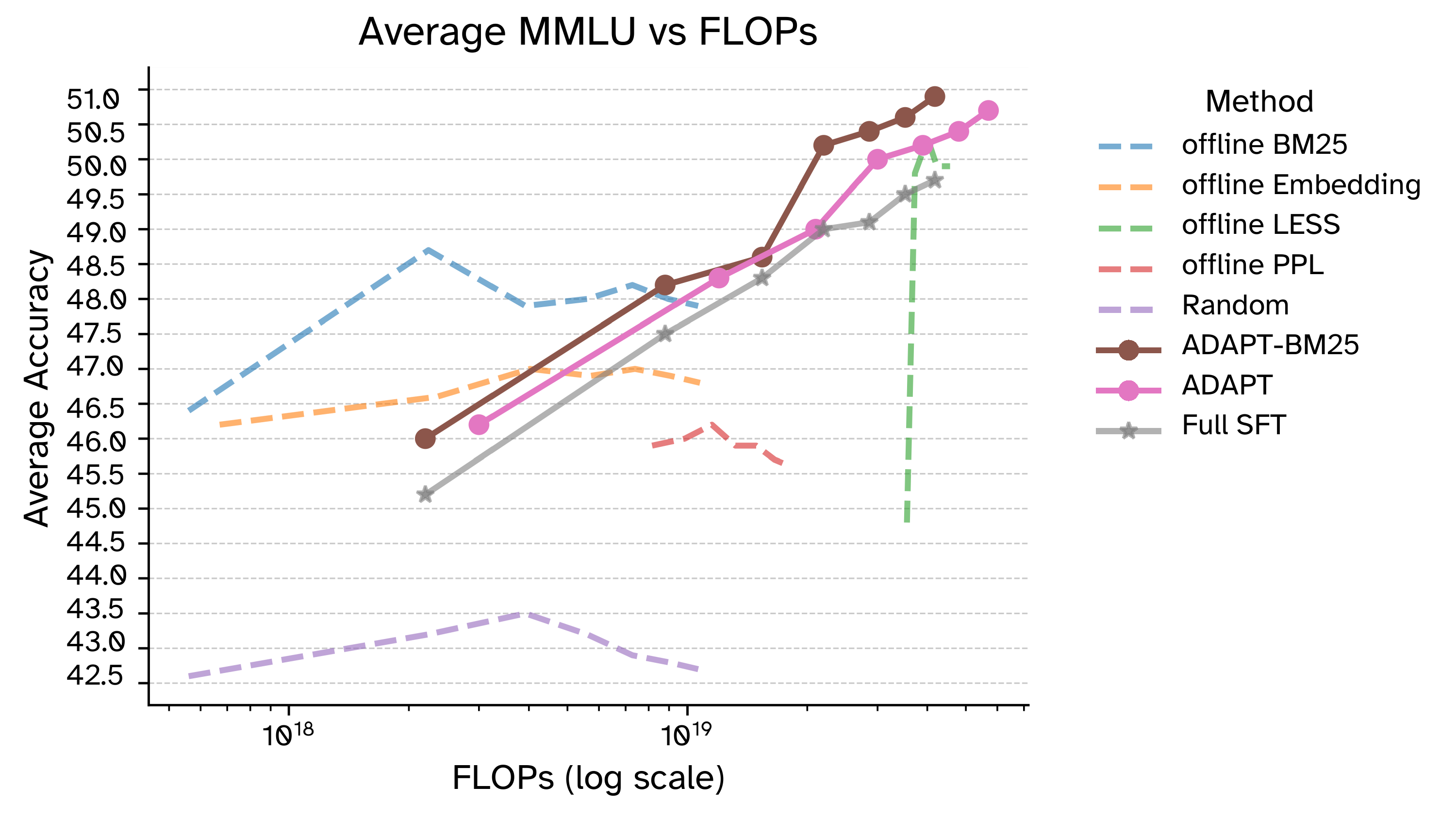}}%
      {\caption{In-domain evaluation: MMLU validation set $\rightarrow$ MMLU benchmark.}%Subset distribution of selected examples.}
      \label{fig:exp_mmlu}
      }
      \ffigbox{\includegraphics[width=1.1\linewidth]{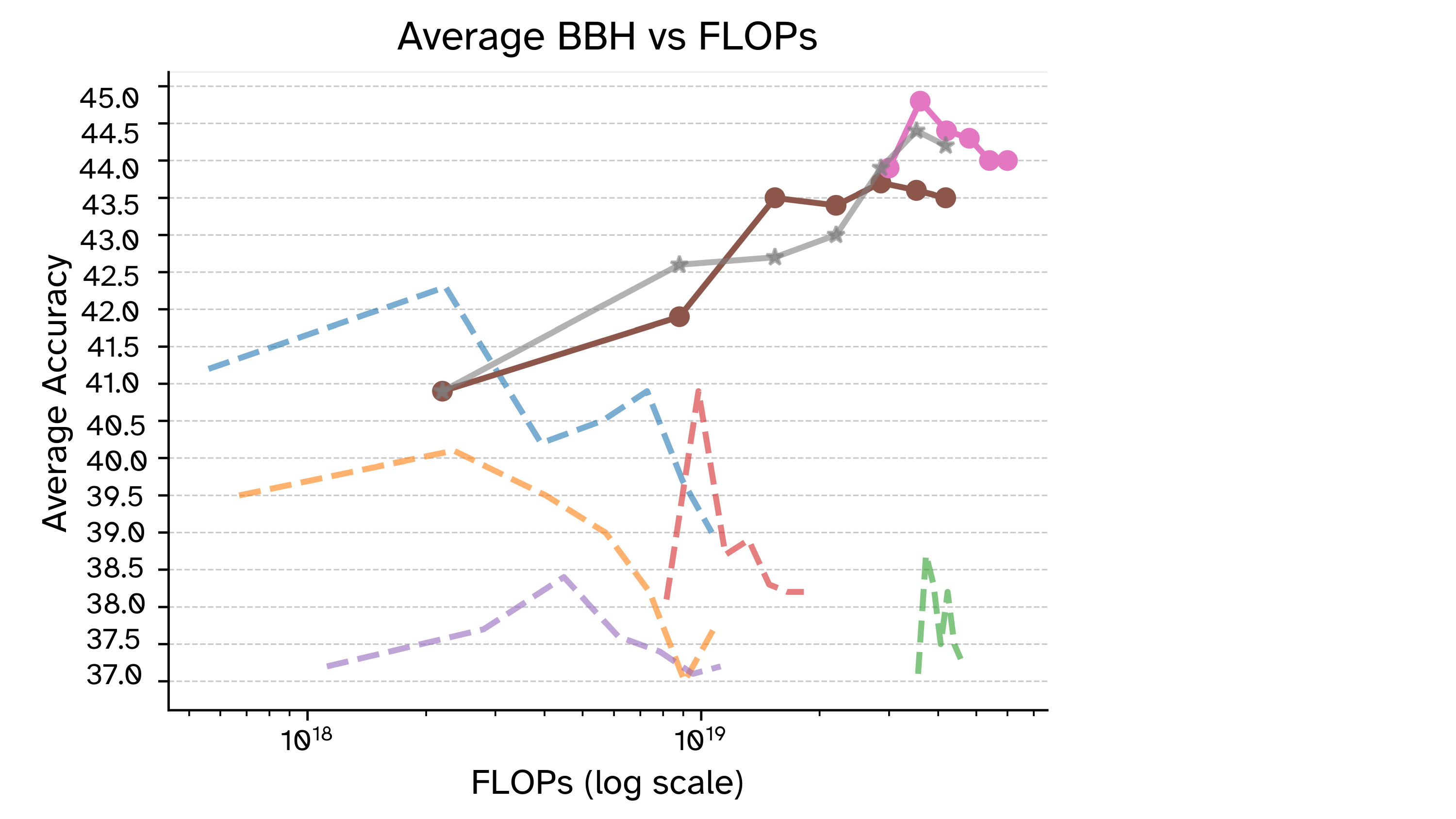}}%
      {\caption{Out-of-domain evaluation: MMLU validation set $\rightarrow$ BBH benchmark.}
      \label{fig:exp_bbh}
      }
    \end{subfloatrow}
  }{%
    \caption{Efficiency-accuracy tradeoff. (a) While offline BM25 is efficient at low FLOPs, it saturates early. ADAPT dominates in the high-performance regime (higher FLOPs), avoiding the early plateauing of heuristics. (b) Crucially, LESS overfits to the in-domain task, showing regression on OOD tasks (BBH), whereas ADAPT consistently improves generalization on the Pareto frontier.%Efficiency--accuracy tradeoff of different offline selection and online reweighting methods under our proposed \textbf{total FLOPs} metric. Offline methods (e.g., \texttt{LESS}) appear competitive under in-domain evaluation, but still weaker than online methods \texttt{ours(bm25), \ours{}}. Also, their advantage vanishes or even reverses in out-of-domain settings. In contrast, online reweighting remains consistently strong across domains.
    }
    \label{fig:exp_sft}
  }
\end{figure}

% We present main results of \ours{} across different benchmarks, and comparison with baseline approaches. we also detailed our estimation of the costs to perform different data selection and reweighting methods. We summarize the findings below.

As shown in Figure~\ref{fig:exp_sft}, our proposed online data reweighting method consistently outperforms existing approaches under comparable computational budgets:

% Our results show that \texttt{BM25\_online} not only avoids additional overhead (since reweighting is integrated into training), but also \emph{outperforms offline methods in out-of-domain evaluation} [x\%]. This demonstrates that online reweighting is not only feasible but superior when judged under a unified and cost-aware evaluation protocol.

\begin{table}[h]
\centering
\small 
\vspace{-1.em}
\begin{tabular}{l c c}
\toprule
Method & MMLU\textit{(val)} - MMLU\textit{(test)} & MMLU\textit{(val)} - BBH\textit{(test)} \\
\midrule
BM25 & 48.7 $\pm$ 0.9 & 42.3 $\pm$ 0.8  \\
Embedding &  47.0 $\pm$ 0.6 & 40.1 $\pm$ 0.5  \\
LESS  &  50.2 $\pm$ 0.5 & 38.7 $\pm$ 1.5 \\
PPL  &  46.2 $\pm$ 1.1 & 40.9 $\pm$ 0.9 \\
Random  & 43.5 $\pm$ 0.3  & 38.4 $\pm$ 1.0 \\
\midrule
Full Dataset SFT  & 49.7 $\pm$ 0.2 & 44.4 $\pm$ 0.3  \\
\rowcolor{gray!15}
\ours{}-BM25  & \textbf{50.9 $\pm$ 0.6} &  \textbf{43.7 $\pm$ 1.2} \\
\rowcolor{gray!15}\textbf{\ours{}} & \textbf{50.7 $\pm$ 0.7} &  \textbf{44.8 $\pm$ 1.3} \\
\bottomrule
\end{tabular}
\caption{\centering Generalization results under different validation-test configurations.}
\label{tab:exp_instruction}
\end{table}

\textbf{Comparison with offline data selection methods.} Across all baselines in both Figure ~\ref{fig:exp_mmlu} and Figure ~\ref{fig:exp_bbh}, our approach consistently lies on the Pareto frontier, demonstrating the best trade-off between computational efficiency and final accuracy. At the same FLOPs, our method achieves substantially higher accuracy than the offline data selection SoTA method LESS. To reach the same performance level, our approach requires significantly fewer FLOPs than LESS. Moreover, after reaching this performance, further training with LESS leads to overfitting and accuracy degradation, whereas online data reweighting continues to exhibit steady performance gains. Notably, our online reweighting method using BM25 as quality signal (\ours{}-BM25) has better generalization than its offline selection method (BM25) with all other configurations kept the same, which demonstrates the advantage of online data curation.  

\textbf{Comparison with full-data fine-tuning.} In Figure~\ref{fig:exp_mmlu} and Table~\ref{tab:exp_instruction}, our method also outperforms full-data fine-tuning under equal FLOPs, with the performance gap widening as training progresses. In Fig.~\ref{fig:exp_bbh} and Table~\ref{tab:exp_instruction}, our reweighting strategy that incorporates model state information (\ours{}) outperforms our fixed-state online reweighting (\ours{}-BM25), highlighting the importance of adapting to the evolving model state to better generalization. 
\textbf{Generalization cross benchmark.}
As shown in Figure~\ref{fig:exp_bbh}, we further validate the generalization capability of models trained with \ours{}, where we use MMLU~\citep{mmlu} as the validation set to select the data, and evaluate the model trained on selected data on the BBH~\citep{bbh} task. 
The results, also reported in Table~\ref{tab:exp_instruction}, show that \ours{} achieves better generalization performance across benchmarks compared to other offline data selection baselines and demonstrates comparable generalization with vanilla full dataset instruction tuning. % It shows that \ours{} can successfully retain original knowledge.
This highlights a key strength of \ours{}: when new downstream tasks or benchmarks emerge, there is no need to repeat the entire data selection and model training pipeline. Over the long term, this substantially reduces FLOPs costs, making \ours{} highly practical for real-world applications.

% \vspace{-1.8em}
\begin{table}[!t]

\setlength{\tabcolsep}{3.4pt}
\centering
\small 
% \vspace{-.5em}
\caption{\textbf{Benchmark performance of \emph{TinyLlama-120M} trained on 50B tokens using Uniform, LinUpper, DoReMi, RegMix, and \ours{}.}}
\vspace{-.5em}
\begin{tabular}{lcccc|cc}
\toprule
\textbf{Tasks} & \textbf{Uniform} & \textbf{LinUpper} & \textbf{DoReMi} & \textbf{RegMix} & \textbf{\ours{}-BM25} & \textbf{\ours{}} \\
\midrule
ARC-C & $17.75_{\tiny \pm 0.00}$ & $18.77_{\tiny \pm 0.24}$ & $18.00_{\tiny \pm 0.84}$ & $18.60_{\tiny \pm 0.06}$ & $18.31_{\tiny \pm 0.55}$ & \textbf{$19.11_{\tiny \pm 0.60}$} \\
\rowcolor{gray!15} ARC-E & $39.60_{\tiny \pm 0.10}$ & $38.85_{\tiny \pm 0.12}$ & $40.70_{\tiny \pm 0.87}$ & \textbf{$41.04_{\tiny \pm 0.21}$} & $39.44_{\tiny \pm 0.60}$ & $39.06_{\tiny \pm 0.54}$\\
COPA & $63.00_{\tiny \pm 0.00}$ & $60.00_{\tiny \pm 1.72}$ & \textbf{$66.00_{\tiny \pm 1.95}$} & $61.00_{\tiny \pm 1.5}$ & $63.20_{\tiny \pm 0.40}$ & $64.00_{\tiny \pm 1.32}$ \\
\rowcolor{gray!15} HellaSwag & \textbf{$28.39_{\tiny \pm 0.00}$} & $27.90_{\tiny \pm 0.13}$ & $28.03_{\tiny \pm 0.03}$ & $27.63_{\tiny \pm 0.03}$ & $27.96_{\tiny \pm 0.07}$ & $28.11_{\tiny \pm 0.00}$ \\
 Lambada-O & \textbf{$24.68_{\tiny \pm 0.64}$} & $23.64_{\tiny \pm 0.28}$ & $22.38_{\tiny \pm 0.73}$ & $24.82_{\tiny \pm 0.94}$ & $26.19_{\tiny \pm 0.30}$ & $24.63_{\tiny \pm 0.53}$ \\
Lambada-S & $16.98_{\tiny \pm 0.81}$ & $16.79_{\tiny \pm 0.91}$ & $16.30_{\tiny \pm 1.05}$ & $18.30_{\tiny \pm 0.14}$ & $18.43_{\tiny \pm 0.33}$ &\textbf{$18.07_{\tiny \pm 0.79}$} \\
\rowcolor{gray!15} LogiQA & $20.43_{\tiny \pm 0.54}$ & $21.35_{\tiny \pm 0.03}$ & $21.51_{\tiny \pm 0.11}$ & $20.89_{\tiny \pm 0.07}$  & $20.52_{\tiny \pm 0.69}$ & \textbf{$21.66_{\tiny \pm 0.69}$} \\
MultiRC & $56.68_{\tiny \pm 0.00}$ & $50.74_{\tiny \pm 0.29}$ & $48.99_{\tiny \pm 2.57}$ & \textbf{$56.70_{\tiny \pm 0.93}$} & $56.53_{\tiny \pm 0.44}$ & $55.67_{\tiny \pm 1.11}$ \\
\rowcolor{gray!15} OpenBookQA & $15.80_{\tiny \pm 0.40}$ & $14.00_{\tiny \pm 0.18}$ & $15.20_{\tiny \pm 0.60}$ & \textbf{$18.20_{\tiny \pm 0.50}$} & $15.40_{\tiny \pm 0.72}$ & $14.20_{\tiny \pm 0.43}$ \\
PIQA & $60.55_{\tiny \pm 0.29}$ & $60.83_{\tiny \pm 0.45}$ & $59.90_{\tiny \pm 0.39}$ & $58.98_{\tiny \pm 0.29}$ & $61.00_{\tiny \pm 0.19}$ & \textbf{$61.48_{\tiny \pm 0.23}$} \\
\rowcolor{gray!15} QQP & 36.88$_{\tiny \pm 0.00}$ & \textbf{$37.85_{\tiny \pm 1.21}$} & $36.84_{\tiny \pm 0.04}$ & $36.83_{\tiny \pm 0.00}$ & $36.82_{\tiny \pm 0.00}$ & $36.81_{\tiny \pm 0.00}$ \\
RACE & \textbf{$27.85_{\tiny \pm 0.14}$} & $26.41_{\tiny \pm 0.18}$ & $26.41_{\tiny \pm 0.03}$ & \textbf{$27.85_{\tiny \pm 0.00}$} & $27.39_{\tiny \pm 0.35}$ & $26.60_{\tiny \pm 0.49}$ \\ 
SciQ & $71.00_{\tiny \pm 1.00}$ & $70.70_{\tiny \pm 1.34}$ & \textbf{$72.90_{\tiny \pm 0.23}$} & $71.20_{\tiny \pm 0.50}$ & $72.48_{\tiny \pm 0.44}$ & $72.50_{\tiny \pm 1.37}$ \\
SocialIQA & $36.90_{\tiny \pm 0.03}$ & $36.85_{\tiny \pm 0.16}$ & $36.95_{\tiny \pm 0.07}$ & \textbf{$37.41_{\tiny \pm 0.46}$} & $36.52_{\tiny \pm 0.47}$ & $37.05_{\tiny \pm 0.30}$ \\
\rowcolor{gray!15} WinoGrande & $50.75_{\tiny \pm 0.17}$ & $50.83_{\tiny \pm 0.09}$ & $49.64_{\tiny \pm 0.79}$ & $50.12_{\tiny \pm 0.72}$ & $50.61_{\tiny \pm 1.25}$ & \textbf{$50.99_{\tiny \pm 0.00}$} \\
\midrule
Average (All) ($\uparrow$) & $37.81_{\tiny \pm 0.13}$ & $37.03_{\tiny \pm 0.12}$ & $37.32_{\tiny \pm 0.15}$ & $37.97_{\tiny \pm 0.02}$ & $38.05_{\tiny \pm 0.21}$ & \textbf{$38.00_{\tiny \pm 0.22}$} \\
% \rowcolor{gray!15} \todo{Average (Seen) ($\uparrow$)} & $41.71_{\tiny \pm 0.10}$ & $40.31_{\tiny \pm 0.10}$ & $40.88_{\tiny \pm 0.10}$ & $40.92_{\tiny \pm 0.10}$ & \textbf{$41.33_{\tiny \pm 0.10}$} & \textbf{$40.78_{\tiny \pm 0.10}$} \\
\rowcolor{gray!15} Average (Unseen) ($\uparrow$) & $31.98_{\tiny \pm 0.09}$ & $30.56_{\tiny \pm 0.28}$ & $31.95_{\tiny \pm 0.17}$ & $32.46_{\tiny \pm 	0.39}$ & \textbf{$33.49_{\tiny \pm 0.37}$} & \textbf{$33.73_{\tiny \pm 0.39}$}\\
\midrule
FLOPs overhead ($\downarrow$) & 0 & 0 & $4.92 \times 10^{19}$ & $3.072 \times 10^{18}$ & {$\ll$
 $1.0 \times 10^{14}$} & $\ll 1.1 \times 10^{15}$ \\
\bottomrule
\end{tabular}
\vspace{-0.1cm}
\label{tab:tinyllama_120M_50B}
\vspace{-0.1cm}
\end{table}
\vspace{-0.1cm}
\vspace{-0.1cm}

\begin{figure}[htbp]
  \centering
  \floatsetup{justification=centering}
  \ffigbox[\FBwidth]{%
    \begin{subfloatrow}[2]
      \ffigbox{\includegraphics[width=1.0\linewidth]{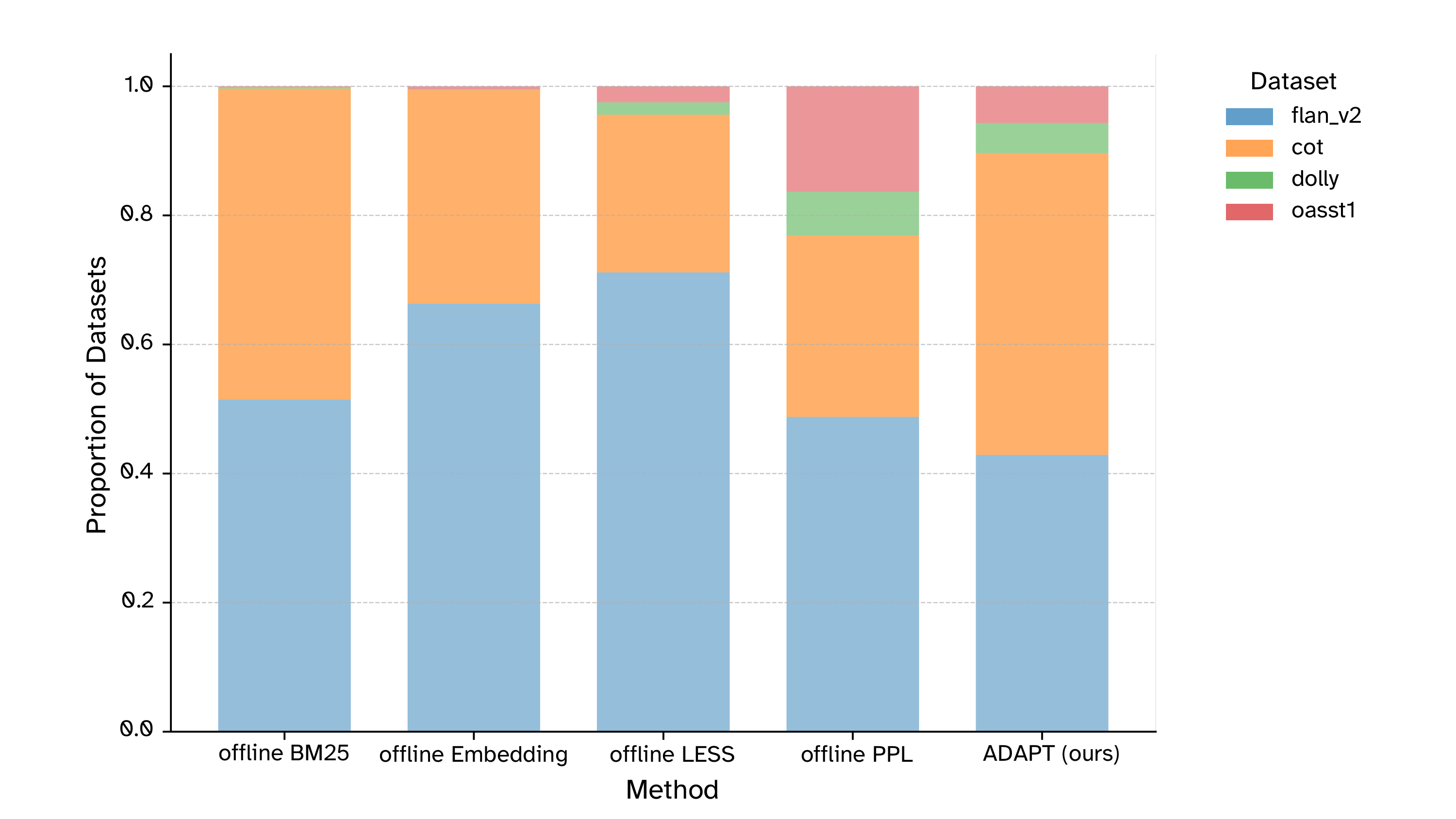}}%
      {\caption{Distribution of effective selected examples}
      \label{fig:mixing}
      }
      \ffigbox{\includegraphics[width=1.0\linewidth]{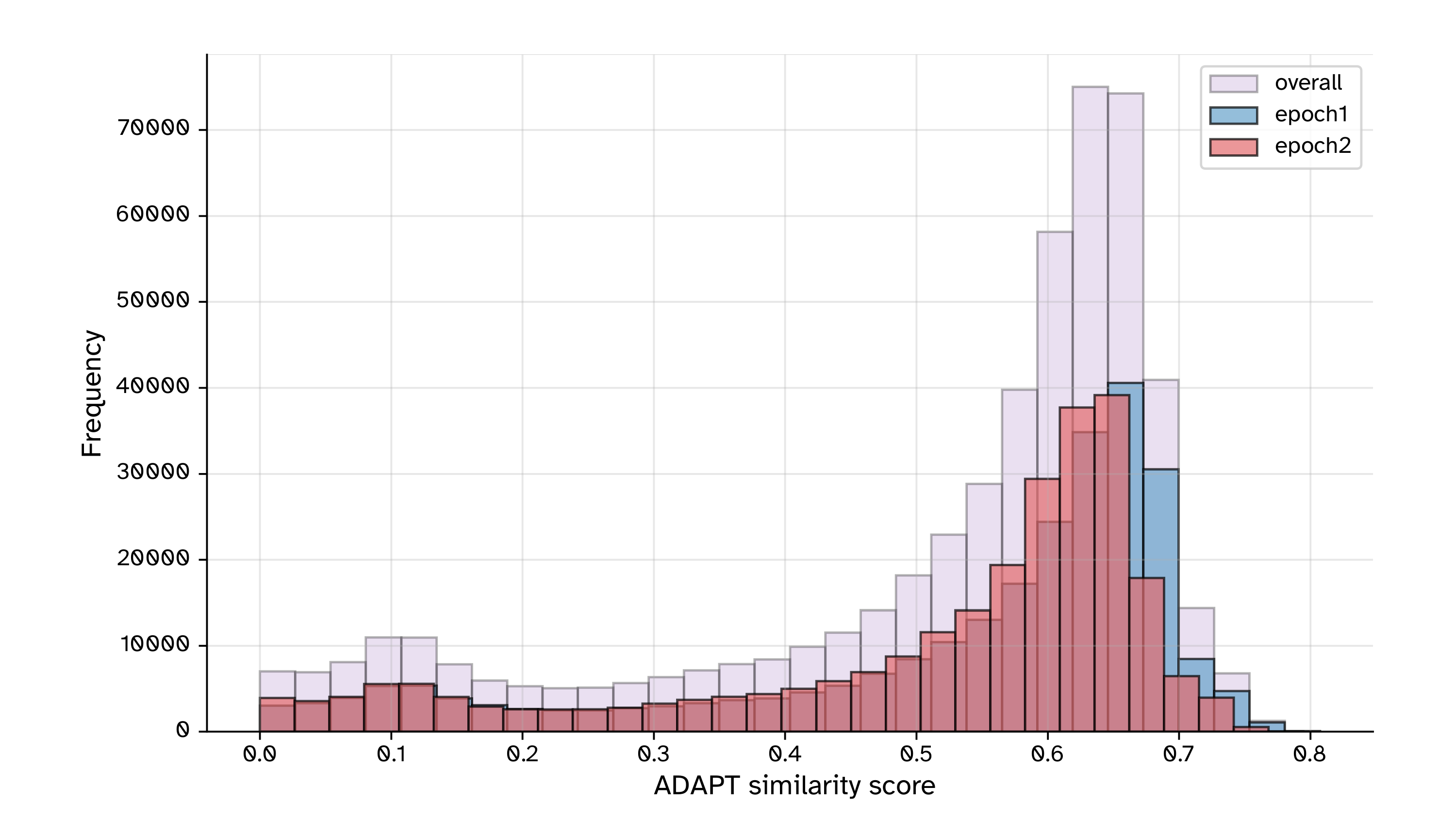}}%
      {\caption{Similarity distribution for different epochs}
      \label{fig:range}}
    \end{subfloatrow}
  }{%
    \caption{A summary of the distribution of effective selected samples and similarity scores. Both differences and changes show the proposed method's ability on capturing data features and adaptive learning.}
    % \label{fig:fitfalls}
  }
  \vspace{-0.8em}
\end{figure}

\paragraph{Data mixing and curriculum. }
We present the distribution of the effective selected examples for different methods in Figure ~\ref{fig:mixing}. Different approaches end up selecting very different amounts of data from each training subset. It demonstrates that online reweighting can also help with deciding proper data mixture. We also calculate the \emph{effective proportion of used data}, by summing up weights of all data together, with the result of $0.501$ in our experimental setting. In practice, this number would automatically adapt according to the overall quality of the training corpus we are selecting from.

We also presented Figure ~\ref{fig:range}. The shift in similarity distributions from epoch 1 to epoch 2 reflects a transition from collapsed, overly homogeneous representations toward more diverse and fine-grained embeddings. This diversification enhances generalization by improving the model's ability to capture subtle distinctions and reducing over-reliance on coarse features. Notably, this process can be viewed as an instance of implicit curriculum learning: the model first clusters samples based on coarse, ``easier'' patterns and progressively moves toward harder, fine-grained discrimination, thereby mirroring the principles of curriculum learning without explicit scheduling.

\vspace{-.5em}
\vspace{-0.1cm}
\begin{wraptable}{r}{0.55\textwidth}  % r = right side, l = left side
\setlength{\tabcolsep}{3.4pt}
\renewcommand{\arraystretch}{1.05}
\vspace{-0.2cm}
\caption{\textbf{Benchmark performance of \emph{TinyLlama-120M} trained on 100B tokens using Uniform and \ours{}.}}
\begin{tabular}{lcc}
\toprule
\small 
\textbf{Tasks} & \textbf{Uniform} & \textbf{\ours{}} \\
\midrule
ARC-C & 17.75 & \textbf{18.86} \\
\rowcolor{gray!15} ARC-E & \textbf{40.66} & 40.45 \\
COPA & \textbf{63.00} & 61.00 \\
\rowcolor{gray!15} HellaSwag & 28.35 & \textbf{28.75} \\
Lambada-O & 25.13 & \textbf{25.91} \\
Lambada-S & 18.47 & \textbf{20.14} \\
\rowcolor{gray!15} LogiQA & 20.28 & \textbf{21.66} \\
MultiRC & 55.98 & \textbf{56.44} \\
\rowcolor{gray!15} OpenBookQA & \textbf{16.00} & 15.00 \\
PIQA & 61.48 & \textbf{61.97} \\
\rowcolor{gray!15} QQP & 36.81 & \textbf{36.82} \\
RACE & 27.37 & \textbf{27.46} \\
SciQ & 71.50 & \textbf{72.40} \\
SocialIQA & \textbf{35.77} & 37.05 \\
\rowcolor{gray!15} WinoGrande & 51.14 & \textbf{51.54} \\
\midrule
Average (All) ($\uparrow$) & 37.98 & \textbf{38.36} \\
% \todo{\rowcolor{gray!15} Average (Seen) ($\uparrow$) } & 41.99 & \textbf{42.49} \\
\rowcolor{gray!15} Average (Unseen) ($\uparrow$) & 32.51 & \textbf{32.67} \\
\bottomrule
\end{tabular}
\label{tab:tinyllama_120M_100B}
\vspace{-0.2cm}
\end{wraptable}
\vspace{-0.2cm}

\subsection{Pretraining results}
% \subsection{Online data reweighting also helps in large-scale pretrianing~(\textbf{RQ4})}\label{sec:eval:alignment}

% \input{tables/pretraining_results_full}

% \textbf{Performance on commonsense reasoning tasks.}~~~~
% As shown in Table~\ref{tab:tinyllama_120M_50B}, %selection methods based on a target domain such as DSIR, D4, and QuRating-W improve performance on reading comprehension tasks like ARC-c and OBQA, but suffer significant declines on physical world knowledge tasks especially HellaSwag and PIQA (highlighted in blue), revealing the diversity dilemma. 
%
% In contrast, Doremi, D4, QuRating-A, and our DiSF, which select samples from multiple text domains, achieve competitive results on OBQA and ARC-c while significantly improving overall performance across the remaining tasks.
%
% This highlights the critical importance of diversity in pre-training LLMs.
%
% Notably, 
As shown in Table~\ref{tab:tinyllama_120M_50B}, \ours{} outperforms all baselines in terms of average performance, achieving a 0.19\% improvement over Uniform, 0.68\% over DoReMi, and a 0.97\% over LinUpper. RegMix reaches comparable but slightly lower accuracy on average. Interestingly, LinUpper performs worse than Uniform sampling, suggesting that naive sample-level adjustments may not generalize well. Compared to LinUpper, which conducts normalization and calculates sample-level weights for each batch, our global sample-level reweighting solution demonstrates superior performance in handling unknown ratios of mixed-quality corpora.

To examine the effect of larger training budgets, we extend training to 100B tokens. As shown in Table~\ref{tab:tinyllama_120M_100B}, \ours{} achieves higher benchmark performance than Uniform on 11 out of 15 downstream tasks, with an average improvement of 0.38\%. Table~\ref{tab:train_budget} further shows that \ours{} consistently yields lower validation perplexity than Uniform at both 50B and 100B training budgets. These findings demonstrate that \ours{} improves both validation perplexity and downstream task performance consistently across compute scales.

\section{Conclusion and Discussion }
\vspace{-.5em}

Our work unifies data selection, mixing, and reweighting under a FLOPs-aware framework. We show that offline pipelines often incur significant cost via data preprocessing and faces issues with generalization. In contrast, online reweighting integrates seamlessly into training, adapts to model state, and improves efficiency without affecting effective data size.

% \todo{TODO} Data \emph{selection} (hard masks), \emph{mixing} (sampling or quotas), and \emph{balancing} (loss/gradient scaling) are \textbf{operational variants of the same reweighting principle}. Our discussion focuses on their computational implications—i.e., cost-aware training efficiency \eqref{eq:total-flops} and adaptivity to the training state \eqref{eq:online}.
% In this paper, we presented \ours{}, a meta-learning approach that efficiently selects and weights training samples based on gradient alignment with validation examples. 
% Empirically, \ours{} consistently outperformed offline and online baselines across both instruction tuning and pretraining. It achieved lower perplexity, stronger downstream accuracy, and better generalization under fixed compute budgets. These results highlight the risks of static data mixtures and the benefits of adaptive, end-to-end data curation.
Empirically, we demonstrate that \ours{} consistently outperforms both offline and online baselines across instruction tuning and pretraining setups. In instruction tuning, \ours{} not only achieves higher accuracy under equal FLOPs but also exhibits stronger cross-benchmark generalization (e.g., from MMLU to BBH), mitigating the brittleness of offline methods. In pretraining, \ours{} improves both downstream task accuracy and validation perplexity under 50B and 100B token budgets, underscoring its robustness across compute scales. We further discuss the implications and outline future directions in Sec.~\ref{sec:extended_discussion}.

\section*{Ethics and Reproducibility Statements}

We adhere to the ICLR Code of Ethics. This paper focuses on comparing methodologies for data curation. To ensure fairness, we evaluate online data reweighting and offline data mixing methods under a FLOPs-equivalent setting. We also rigorously test their performance in both instruction fine-tuning and pretraining scenarios to provide a comprehensive and balanced comparison.
During the preparation of this manuscript, we utilized large language models (LLMs) to assist with grammar correction and refinement of the writing.
We provide all necessary details to ensure reproducibility of our work. The theoretical justification for our FLOPs-equivalent data curation formalization is presented in Section~\ref{sec:unification_formalization}, with detailed FLOP calculations in Appendix~\ref{Appendix-DS-FLOPS}. Implementation details and training protocols are provided in Section~\ref{sec:exp_design}, Appendix~\ref{subsec:baselines}, Appendix~\ref{app:pretrain_details}, and Appendix~\ref{app:lora}, while descriptions of the training data are given in Appendix~\ref{app:train_data}.

% training methodology to enhance MoE upcycling, however, the model itself incorporates an LLM that may perpetuate biases present in the training data, potentially affecting fairness and reliability. Therefore, we recommend adhering to standard ethical guidelines for the use of LLMs to mitigate these risks.

% During the preparation of this manuscript, we utilized large language models (LLMs) to assist with grammar correction and refinement of the writing.

% In this paper, we provide all the necessary details to ensure the reproducibility of the presented method. We provide the theoretical justification of the method in Section 2 and Appendix A.1, implementation details and training protocoles in Section 3.1, Appendix A.4 and Appendix A.6, and data description in Appendix A.5.

% % \section*{The Use of LLMs in Writing}

% %We used LLM, namely \textsc{OpenAI-GPT5}, to polish the writing of this manuscript. No other generative AI functionality is used in the writing of this submission.  

% %\section*{Reproducibility}

\section*{Acknowledgement}
We would like to thank Zhihao Zhan, Ziqing Fan, Jonathan Hayase, Guilherme Penedo, Hynek Kydl{\'\i}{\v{c}}ek and Loubna Ben allal for many stimulating and helpful discussions. This research was supported by the following entities: The Royal Academy of Engineering via DANTE (a RAEng Chair); the European Research Council, specifically the REDIAL project; SPRIND under the composite learning challenge; Google through a Google Academic Research Award. Yihong acknowledges funding from the Horizon Europe grant No. 101213369 (DVPS).

\bibliography{iclr2026_conference}
\bibliographystyle{iclr2026_conference}

\clearpage
\appendix
\addtocontents{toc}{\protect\setcounter{tocdepth}{2}}

\section*{Appendix}
\tableofcontents
\clearpage
% We provide further details in App.~\ref{app:flops_details}.

\section{Implication and Future Work.} 
\label{sec:extended_discussion}
\textbf{Implication.} \ours{} could be effectively used for getting better control under constrained budgets (both \textit{data budget} and \textit{compute budget}). Our empirical observation provides a principled view and reference for automatically allocating computational resources to the most valuable samples while maintaining end-to-end efficiency. 
Additionally, we expect our method to perform even better where the data quality is often low, which is the common case in practical setups, where data is crude and is not ideal for direct use for LLM training, sources of generally low quality, and biases inherent to the distribution of content on the web.
% We aim to develop principled and effective techniques to fundamentally revamp pretraining dataset curation. 

\textbf{Future Work.} We didn't explicitly discuss and evaluate deduplication, which is a key step for data preprocessing. For example, we will treat deduplication as reweighting related documents in accordance with their frequency. Based on this framework, we will explore different trade-offs inherent in each stage, such as quality-aware deduplication where high-quality documents are allowed to be duplicated more than once. ~\nocite{jiang2023hexgen,jiang2025demystifying,jiang2025hexgen, yan2024hexiscale, yan2025fsa, jiang2025cascadiaefficientcascadeserving}

In future work, we aim to explore ADAPT for safety-oriented data curation~\citep{gal2025customizable} and to study how dynamic data reweighting influences memorization and internal knowledge structure formation in large language models~\citep{chen2025structure}. In particular, combining ADAPT with interpretability techniques such as influence functions~\citep{basu2020influence} and Jet Expansions~\citep{chen2026jet} may enable us to trace how safety-critical data shapes learned representations and to examine whether models preserve underlying data distributions under dynamic reweighting. More broadly, we view ADAPT as a mechanism for dynamically modulating what information is reinforced during training. Such \textit{data-centric} control of model memory opens avenues not only for improving model quality and safety but also for gaining insights into LLM internal knowledge formation when paired with model-side interpretability methods.
\nocite{macfarlane2025instilling, zhao2025breakingphysicallinguisticborders, zhao2025cluescollaborativehighqualitydata, zhao2024attacksthirdpartyapislarge}

\section{Instruction Tuning Baselines}
\label{subsec:baselines}
We compare \ours{} with the following baselines: \textbf{1) Random Selection}: We randomly sample data from the instruction tuning dataset. \textbf{2) BM25}~\citep{robertson2009probabilistic}: We assign weights to training samples based on textual statistical features (i.e., TF-IDF), and select the top $k$ data points with the highest scores. %\textbf{3) \todo{embedding} DSIR}~\citep{xie2023data}: Weassign weights to training samples based on $n$-gram features, and resample $k$ data points according to the importance weights. 
%
%We also compare to \textbf{RDS} (Representation-based Data Selection)~\citep{zhang2018unreasonable, hanawa2020evaluation}, which uses model's hidden representations as features for data selection. For fair comparison, we compute the similarity score using~\Cref{eq:matching} but replace the gradient features with the final layer representations of the last token of each sequence. These features are $2048$-dimensional in \llamasmall{}. % Another intuitive baseline we consider is to directly use the validation data $\dval$ for training, but this drastically degrades performance (\Cref{app:valid}).
% \textbf{DSIR}~\citep{xie2023data}
\textbf{4) PPL}~\citep{yin2024computeconstraineddataselection,marion2023moreinvestigatingdatapruning,ankner2024perplexedperplexityperplexitybaseddata}: We compute the loss of each training sample on our original base model and use it as its score. % \textbf{5) IFD}~\citet{li-etal-2024-quantity}: We first training a model on representative samples from the dataset, and then scoring data points using the ratio of the answer loss given the question to the loss of the answer on its own. %We compute the IFD score for all points $d \in D$. We use the codebase provided by the authors.\footnote{\url{https://github.com/tianyi-lab/Cherry_LLM}} When selecting data from the 5.8M \modelname 2 unfiltered pool, we use the same model trained on the 200k-size pool, as the smaller pool is simply a subsampled version of the larger one.
\textbf{6) LESS}~\citet{xia2024less}: We train LoRAs on a random subset of the data, and then selecting data by computing the gradient-based influence of each training sample to validation samples. % \textbf{7) LinUpper}~\citep{sow2025dynamic}: linear upper-bound strategy where the weight is proportional to the normalized loss but is capped at a predefined $\alpha$ value, ensuring that outliers do not dominate the training process. 

\section{Data-Selection FLOPs}
\label{Appendix-DS-FLOPS}
% For the FLOPs estimates used throughout the paper, we follow \citet{kaplan2020scalinglawsneurallanguage} in estimating the compute cost of a training step as roughly $6N$ FLOPs per token processed, where $N$ is the parameter count of the model (roughly 7B). \citet{kaplan2020scalinglawsneurallanguage} notes that the forward pass is roughly half the cost of the backward pass, giving us an estimate of $2N$ FLOPs per token when processing samples. We use a rough estimate of 2,048 tokens per sample, since during training and selection we truncate all samples to be at most this length. Note we fully-finetune models for two epochs in all setups. Let $N$ be the model size, $P$ be the size of the data pool (in number of samples), and $D$ the number of samples selected to train on. Based on this, the cost for each method is estimated as follows:
To estimate computational costs throughout our paper, we adopt the methodology of \citet{kaplan2020scalinglawsneurallanguage}, which approximates the training step computation as approximately $6N$ FLOPs per processed token, where $N$ represents the model's parameter count (approximately 7B). According to \citet{kaplan2020scalinglawsneurallanguage}, the forward pass consumes roughly half the computational resources of the backward pass, yielding an estimate of $2N$ FLOPs per token during sample processing. We employ an approximation of 2,048 tokens per sample, as we limit all samples to this maximum length during both the training and selection phases. Note that in all experimental configurations, we conduct full fine-tuning of models over two complete epochs. If we define $N$ as model size, $P$ as the data pool magnitude (measured in sample quantity), and $D$ as the number of samples chosen for training, we can calculate the computational expense for each methodology as follows:
\begin{enumerate}
    \item \textbf{Random Selection}: $2k \times 6N \times D \times E$
    \item \textbf{BM25}: $2k \times 6N \times D \times E$
    \item \textbf{Embedding}: $2k \times 2N'\times P + 2k * 6N \times D \times E$ (embed model $N'$)
%     \item \textbf{DSIR}: $2 \times 2048 \times c_{\mathrm{hash}} \times P +  2 \times 2048 \times 6N \times D$
    \item \textbf{PPL}: $2k \times 2NP + 2k \times 6N \times D \times E$
    % \item \textbf{IFD}: $270679 \times 2048 \times 2N + 1000 \times 2048 \times 6ND + 2 \times 2048 \times 2NP + 2 \times 2048 \times 6ND$ (We train the initial model used to compute IFD scores on 1000 samples selected from the 200k data pool.)
    \item \textbf{LESS}: $1.53 \times 2k \times 6N \times P + 2k \times 6N \times D \times E$ (LESS computes gradients for three checkpoints over the entire pool.)
%     \item \textbf{LinUpper}: $2048 \times 2N \times P + 2 \times 2048 \times 6N \times P$ (LinUpper first performs forward passes on the entire data pool to compute linear upper bounds and assign weights, then uses these weights to fine-tune on the entire data pool for two epochs.)
    %\item \textbf{\ours{}-BM25}: % $ 2048 \times + 2 \times 2048 \times 6ND$ 
    %$2k \times 6N\times P \times E$
    %\item \textbf{\ours{}}: % $ 2048 \times + 2 \times 2048 \times 6ND$ 
    %$2k \times 8N\times P \times E$
    %2 \times 1000 \times 2048 \times 8N + 2048 \times 6N \times D $ (\ours{} requires the forward and backward computation for both training set and meta data. ) %+ M \times D \times L \times (4r + 10 + 10Kr)$    
    %\item %\textbf{Embedding}: $2 * 2048 * 2NP + 2 * 2048 * 6ND$
    % \item % \textbf{RDS+}: $2 * 2048 * 2NP + 2 * 2048 * 6ND$
\end{enumerate}

\section{Training Datasets in Instruction Tuning}
\label{app:train_data}
Table \ref{tab:train_data} contains information about the training sets used in instruction tuning.
% For training, we utilize the same four processed datasets as referenced in \citep{wang2023far}. Each of these datasets is either annotated or authored by humans. Further details about these datasets can be found in \ref{tab:train_data}. The \flan{} and \cott{} datasets are derived from existing NLP datasets, while \dolly{} and \oasst{} consist of open-ended generation examples featuring answers written by humans. These datasets vary significantly in format, sequence length, and underlying tasks, reflecting the heterogeneity of instruction-tuning datasets. We adopt the following 'Tulu' format from the study by~\citet{wang2023far} to standardize the format of these datasets.

\begin{table*}[h]
    \centering
    \resizebox{0.98\textwidth}{!}{%
    \begin{tabular}{lrlccc}
        \toprule
        
    \textbf{Dataset}                & \textbf{\# Instance} & \textbf{Sourced from}                              & \textbf{\# Rounds} & \textbf{Prompt Len.} & \textbf{Completion Len.} \\ \cmidrule(r){1-6}
    \flan & 100,000                         & NLP datasets and human-written instructions & 1                             & 355.7                          & 31.2                               \\
    \cott   & 100,000  & NLP datasets and human-written CoTs         & 1  & 266   & 53.2   \\
    \dolly  & 15,011                          & Human-written from scratch                & 1                             & 118.1                          & 91.3                               \\ 
    \oasst & 55,668                          & Human-written from scratch                & 1.6                           & 34.8                           & 212.5     \\ \bottomrule                        
    \end{tabular}}
    \caption{Details of training dataset from \citet{wang2023far}. Len. is short 
    for token length. }
    \label{tab:train_data}
\end{table*}

%\section{Baseline Details} \label{app:baselines}

%\paragraph{LinUpper} The functional form is $s_i := \min\{h_i + \alpha, \alpha\}$. 
%
%By dynamically adjusting the importance of individual samples based on their loss values, our approach overcomes the limitations of traditional uniform averaging methods and adds a new dimension to existing domain-level reweighting methods by incorporating more fine-grained sample-level dynamics. Through extensive theoretical and empirical validation, we demonstrate that down-weighting low-loss samples accelerates convergence and improves performance. The experiments show that our proposed \texttt{LinUpper} strategy consistently outperforms uniform sampling on common LLM reasoning and commonsense benchmarks. 

\section{Evaluation Datasets in Instruction Tuning}
\label{app:eval_data}
Table \ref{tab:eval_dataset} contains detailed statistics of the evaluation datasets used in instruction finetuning.

\begin{table}[h]
    \centering
    \resizebox{0.66\textwidth}{!}{%
    \begin{tabular}{lrrrrl}
    \toprule
    \textbf{Dataset}                & \textbf{\# Shot} & \textbf{\# Tasks} & $\vert \boldsymbol{\mathcal{D}}_{\boldsymbol{\mathrm{val}}} \vert$ & $\vert \boldsymbol{\mathcal{D}}_{\boldsymbol{\mathrm{test}}} \vert$ & \textbf{Answer Type} \\ \midrule
    \mmlu   &5 & 57   & 285    & 18,721        & Letter options \\ 
    %\tydiqa & 1 & 9   & 9      & 1,713         & Span \\
    \bbh    & 3 & 23   & 69   & 920          & COT and answer \\ 
    %\tydiqa & 1 & 9   & 9      & 1,713         & Span \\
    \bottomrule 
    \end{tabular}}
    \caption{Statistics of evaluation datasets. The selection of evaluation tasks cover different kinds of answer types.}
    \label{tab:eval_dataset}
\end{table}

\section{Pretraining Details} \label{app:pretrain_details}
We follow all settings in TinyLlama~\citep{tinyllama}. 
The optimizer is AdamW~\citep{adamw}, setting parameters $\beta_1$ at 0.9 and $\beta_2$ at 0.95. 
We adopt the cosine learning rate schedule with a maximum learning rate of 4e-4 and the minimum of 4e-5, the batch size of 2M tokens, the weight decay of 0.1, and the gradient clipping threshold of 1. 
%
% The training budgets are 10B and 50B tokens, with 1.5\% selection budget of SlimPajama's training files.
%
% Note that, we choose to report performance with a 1.5\% selection budget, since it achieves comparable performance compared to Full Data pre-training under 50B pre-training budget on TinyLlama 1.1B.
%
% Unless otherwise specified, the selection scale $b$ is set to 1024.
%
% See Appendix~\ref{app:model} for more details.

\section{Instruction Tuning Details}\label{app:lora}
All experiments were conducted with parameter-efficient finetuning method LoRA~\citep{hu2021lora}. For the LoRA adapter, we specified a rank of 128, an $\alpha$ value of 512, and a dropout rate of 0.1 and applied it across all attention matrices. Adding the LoRA adapter introduces minimal FLOPs overhead during training---having no impact on our FLOPS analysis---and mainly reduce memory requirements for more accessible training.

\section{Pretraining Perplexity}

In Table \ref{tab:train_budget}, we evaluate validation perplexity on SlimPajama for TinyLlama-120M under two compute budgets (50B and 100B tokens). Across both settings, \ours{} achieves consistently lower perplexity than the Uniform baseline, highlighting its effectiveness in improving data efficiency.

\begin{table}[h]
\setlength{\tabcolsep}{6pt}
\centering
\caption{\textbf{Validation perplexity of TinyLlama-120M on SlimPajama under different training budgets.}}
\begin{tabular}{lcc}
\toprule
\textbf{Method} & \textbf{50B Train Budget} & \textbf{100B Train Budget} \\
\midrule
Uniform & 17.15 & 15.94 \\
\rowcolor{gray!15} \ours{}    & 16.55 & 15.36 \\
\bottomrule
\end{tabular}
\label{tab:train_budget}
\end{table}

\section{Pseudo Code}

\SetKw{KwOr}{or}
\begin{algorithm}[H]
\SetAlgoLined
\DontPrintSemicolon
\SetKwInOut{Input}{Input}
\SetKwInOut{Output}{Output}
\SetKwComment{Comment}{$\triangleright$}{}

\caption{Online Data Reweighting via Embedding Similarity}
\label{algo:rds-online}

\small

\BlankLine

\Input{Training set $\mathcal{D} = \{(x_i, y_i)\}_{i=1}^N$, validation set $\mathcal{D}_{val} = \{v_j\}_{j=1}^{|\mathcal{D}_{val}|}$, model $f_\theta$, temperature $\tau$, learning rate $\eta$, refresh interval $R$, training steps $T$, small constant $\epsilon > 0$}

\Output{Updated parameters $\theta_T$}

\BlankLine

Initialize model parameters $\theta_0$\;

\BlankLine

\For{$t = 1$ \KwTo $T$}{

    Sample mini-batch $B_t \subset \mathcal{D}$\;

    \BlankLine

    \Comment{Refresh anchor embeddings every $R$ steps via forward pass on current model $\theta_t$ (online update)}

    \If{$t \bmod R = 1$ \KwOr $t = 1$}{

        \ForEach{$v \in \mathcal{D}_{val}$}{

            Forward pass: $(h^{(v)}_1, \ldots, h^{(v)}_{L_v}) \leftarrow f_{\theta_t}(v)$ \;

            Position weights: $w_\ell \leftarrow \ell / \sum_{j=1}^{L_v} j$ for $\ell = 1, \ldots, L_v$\;

            $\phi(v) \leftarrow \sum_{\ell=1}^{L_v} w_\ell h^{(v)}_\ell$\;

            $\phi(v) \leftarrow \phi(v) / \max(\|\phi(v)\|_2, \epsilon)$\;

        }

    }

    \BlankLine

    \Comment{Forward propagation for training samples}

    \ForEach{$(x_i, y_i) \in B_t$}{

        Forward pass: $(\hat{y}_i, h^{(i)}_1, \ldots, h^{(i)}_{L_i}) \leftarrow f_{\theta_t}(x_i)$ \;

        Position weights: $w^{(i)}_\ell \leftarrow \ell / \sum_{j=1}^{L_i} j$ for $\ell = 1, \ldots, L_i$\;

        $\phi(x_i) \leftarrow \sum_{\ell=1}^{L_i} w^{(i)}_\ell h^{(i)}_\ell$\;

        $\phi(x_i) \leftarrow \phi(x_i) / \max(\|\phi(x_i)\|_2, \epsilon)$\;

    }

    \BlankLine

    \Comment{Compute quality scores}

    \ForEach{$i \in B_t$}{

        $s_{ADAPT}(x_i) \leftarrow \frac{1}{|\mathcal{D}_{val}|} \sum_{v \in \mathcal{D}_{val}} \cos(\phi(x_i), \phi(v))$\;

    }

    \BlankLine

    \Comment{Compute per-sample weights}

    \ForEach{$i \in B_t$}{

        $w_t(i) \leftarrow \sigma(s_{ADAPT}(x_i) / \max(\tau, \epsilon))$\;

    }

    \BlankLine

    \Comment{Compute weighted loss}

    $\mathcal{L}_t \leftarrow \sum_{i \in B_t} w_t(i) \ell(\hat{y}_i, y_i)$\;

    \BlankLine

    \Comment{Backward propagation and parameter update}

    $\theta_{t+1} \leftarrow \theta_t - \eta \sum_{i \in B_t} w_t(i) \nabla_{\theta_t} \ell(\hat{y}_i, y_i)$\;

}
\BlankLine

\Return{$\theta_T$}

\end{algorithm}

\section{Qualitative Analysis}

We provide a detailed qualitative comparison in Table~\ref{tab:bm25}, Table~\ref{tab:less} and Table~\ref{tab:adapt}. 
% ==================== MMLU Validation ====================
\newcommand{\mmluval}{%
%\textbf{User:} abstract algebra, college mathematics, elementary mathematics, high school mathematics Answer: \\
%\textbf{Assistant:} xxx \\[4pt]
\textbf{User:} Which expression is equivalent to 5 x 9? A. (5 x 4) x (6 x 5)  B. (5 x 5) + (5 x 4) C. (5 x 5) + (5 x 9) D. (5 x 9) x (6 x 9). Answer: \\
\textbf{Assistant:} B \\[4pt]
\textbf{User:} Let A be the set of all ordered pairs of integers (m, n) such that $7m + 12n = 22$. What is the greatest negative number in the set $B = \{m + n : (m, n) \in A\}$? A. -5 B. -4 C. -3 D. -2 Answer: \\
\textbf{Assistant:} B \\[4pt]
\textbf{User:} Ten students take a biology test and receive the following scores: 45, 55, 50, 70, 65, 80, 40, 90, 70, 85. What is the mean of the students' test scores? A. 55 B. 60 C. 62 D. 65 Answer: \\
\textbf{Assistant:} D %
}

% ==================== BM25 Positive ====================
\newcommand{\bmpos}{%
\textbf{Sample Index:} cot 60736 \\ \textbf{Score:} 0.9403 \\
\textbf{User:}
Question: A multiple choice test consists of 4 questions, and each question has 5 answer choices. In how many ways can the test be completed if every question is unanswered?
Options: (A) 24 (B) 120 (C) 625 (D) 720 (E) 1024
Answer: 5 choices for each of the 4 questions, thus total of $5\times5\times5\times5 = 5^4 = 625$ ways.\\
\textbf{Assistant:} The answer is (C). \\[8pt]

\textbf{Sample Index:} cot 60736 (same entry, continued) \\ \textbf{Score:} 0.9403 \\
\textbf{User:}
Question: In how many different ways can the letters of the word ``CLAIM'' be rearranged?
Options: (A) 110 (B) 120 (C) 125 (D) 130 (E) 140
Answer: The total number of arrangements is $5! = 120$.\\
\textbf{Assistant:} The answer is (B). \\[8pt]

\textbf{Sample Index:} cot 60736 (same entry, continued) \\ \textbf{Score:} 0.9403 \\
\textbf{User:}
Question: A pipe can fill a cistern in 20 minutes whereas the cistern when full can be emptied by a leak in 70 minutes. When both are opened, find when the cistern will be full?
Options: (A) 17 minutes (B) 87 minutes (C) 70 minutes (D) 28 minutes (E) 11 minutes
Answer: $1/20 - 1/70 = (7-2)/140 = 5/140 = 1/28 \Rightarrow$ 28 minutes.\\
\textbf{Assistant:} The answer is (D). \\[8pt]

\textbf{Sample Index:} cot 60736 (same entry, continued) \\ \textbf{Score:} 0.9403 \\
\textbf{User:}
Question: What is the area of a rectangle whose length is twice its width and whose perimeter equals that of a square with area 1?
Options: (A) 1/9 sq units. (B) 4/9 sq units. (C) 8/9 sq units. (D) 7/9 sq units. (E) 6/9 sq units.
Answer: Perimeter of square = 4, so perimeter of rectangle = 4 $\rightarrow$ 2(l + w) = 4 $\rightarrow$ l + w = 2. Since l = 2w $\rightarrow$ 3w = 2 $\rightarrow$ w = 2/3, l = 4/3. Area = (4/3)$\times$(2/3) = 8/9.\\
\textbf{Assistant:} The answer is (C).%
}
% ==================== BM25 Negative ====================
\newcommand{\bmneg}{%
\textbf{Sample Index:} oasst1\_41616 \\ \textbf{Score:} 0.0 \\
\textbf{User:} [Thai] Why is the Lao language similar to Thai?\\
\textbf{Assistant:} [Thai] The Lao and Thai languages are similar because they belong to the Tai--Lao language family in the Indochina region. They share writing systems, grammar, and pronunciation. Their similarity is natural given shared history and culture, as well as long-standing contact and cultural exchange between Laos and Thailand.\\[8pt]

\textbf{Sample Index:} oasst1\_4573 \\ \textbf{Score:} 0.0 \\
\textbf{Question:} Buenos d\'{i}as\\
\textbf{Answer:} Hola buenos d\'{i}as. \textquestiondown En qu\'{e} te puedo ayudar?\\[8pt]

\textbf{Sample Index:} oasst1\_42983 \\ \textbf{Score:} 0.0 \\
\textbf{Question:} [Chinese] First time visiting Shenzhen, please arrange an itinerary for about 3 days.\\
\textbf{Answer:} [Chinese] Shenzhen is a modern city in Guangdong Province, China, known for its high-tech industry, beautiful coastal scenery, and rich shopping and entertainment. Here is a 3-day Shenzhen travel itinerary: [\ldots] (full 3-day itinerary)\\[8pt]

\textbf{Sample Index:} oasst1\_43212 \\ \textbf{Score:} 0.0 \\
\textbf{Question:} [Japanese] I am a cat; I have no name yet.\\
\textbf{Answer:} [Japanese] That is the opening line of Natsume Soseki's novel ``I Am a Cat.''%
}

% ==================== Embedding Positive ====================
\newcommand{\embpos}{%
\textbf{Sample Index:} flan\_v2\_91809 \\ \textbf{Score:} 0.9216 \\
\textbf{User:} [Persian] Question: What type of process is photosynthesis?\\
\textbf{Assistant:} True\\[8pt]

\textbf{Sample Index:} cot\_70063 \\ \textbf{Score:} 0.9224 \\
\textbf{User:} A train 150 meters long completely crosses a 240 meters long bridge in 42 seconds. What is the speed of the train is?\\
Options: (A) 32 (B) 33 (C) 88 (D) 66 (E) 09\\
\textbf{Assistant:} S = (150 + 240)/42 = 390/42 = 33 (B)\\[5pt]
\textbf{Question 2:} A train 420 m long, running with a speed of 63 km/hr will pass a tree in?\\
\textbf{Answer:} 36 sec (A)\\[5pt]
\textbf{Question 3:} In 2005, 30 percent of the math department's 40 students were female [...]\\
\textbf{Answer:} 116% (B)\\[5pt]
\textbf{Question 4:} Seven different objects must be divided among three persons [...]\\
\textbf{Answer:} 196 (D)\\[5pt]
\textbf{Question 5:} John has 5 flavors of chocolate [...]\\
\textbf{Answer:} 31 (B)\\[5pt]
\textbf{Question 6:} The average age of a family of 6 members is 26 years [...]\\
\textbf{Answer:} 16 (C)%
}

% ==================== Embedding Negative ====================
\newcommand{\embneg}{%
\textbf{Sample Index:} dolly\_10969 \quad \textbf{Score:} 0.0 \\[3pt]
\textbf{Question:} How do you make Cincinnati style Venison Chili?\\
\textbf{Answer:} In a 4 to 6 quart pot, bring 1 quart of water to a boil. Add: 2 lbs ground Venison [\ldots] (full venison chili recipe)\\[8pt]

\textbf{Sample Index:} oasst1\_25502 \quad \textbf{Score:} 0.0285 \\[3pt]
\textbf{Question:} How to start fishing?\\
\textbf{Answer:} 1. Obtain the necessary licenses and permits [\ldots] (full beginner fishing guide)\\[8pt]

\textbf{Sample Index:} dolly\_3055 \quad \textbf{Score:} 0.0334 \\[3pt]
\textbf{Question:} What does Tianwen means?\\
\textbf{Answer:} Tianwen means quest for heavenly truth%
}

% ==================== LESS Positive ====================
\newcommand{\lesspos}{%
\textbf{Sample Index:} flan\_v2\_78815 \quad \textbf{Score:} 0.9922 \\[3pt]
\textbf{User:} the compound ratio of 2/3 , 6/7 , 4/3 and 1/8 is given by?\\
Options: a) 2/21 , b) 1/48 , c) 1/98 , d) 1/96 , e) 1/68\\
\textbf{Assistant:} a\\[10pt]

\textbf{Sample Index:} flan\_v2\_40013 \quad \textbf{Score:} 1.0 \\[3pt]
\textbf{User:} Look, this is a photo of Mr. White's family. The man in the photo is Mr. White, the father. The woman is the mother. They have two daughters. The girl behind Mr. White is Mary. She's 14. The girl behind Mrs. White is Jane. She is 11. Mary and Jane are in the same school, but not the same grade. Mary is in Grade2. Jane is in Grade 1. They are good students. How old are Mary and Jane?\\
Options: A) Eleven and thirteen B) Fourteen and eleven C) Eleven D) Thirteen\\
\textbf{Assistant:} B\\[10pt]

\textbf{Sample Index:} flan\_v2\_40832 \quad \textbf{Score:} 0.946 \\[3pt]
\textbf{User:} Please answer the following question: Dogs have an understanding of fair play and become angry if they feel that another dog is getting a better deal, a new study has found. The study looked at how dogs react when a companion is rewarded for the same trick in an unequal way. Friederike Range, a researcher at the University of Vienna in Austria, and her colleagues did a series of experiments with dogs who knew how to respond to the command "give the paw ". The dogs were normally happy to repeatedly give the paw, whether they got a reward or not. Range found that, compared with monkeys, dogs \_.\\
Options: A) care more about whether they are rewarded. B) care less about what they are rewarded with. C) care more about what they are ordered to do. D) care less about who gives them orders.\\
\textbf{Assistant:} B
}

% ==================== LESS Negative ====================
\newcommand{\lessneg}{%
\textbf{Sample Index:} flan\_v2\_81227 \quad \textbf{Score:} 0.0 \\[3pt]
\textbf{User:} Personality: I only sleep 3 hours a day. I have no parents or siblings. I'm incredibly smart. [...] I don't blame you. What is your favorite food? I like well done steak. Candidates 1) Hi, how are you doing? 2) Hi, I am Kathie! I don't like eating fast food. 4) I like Mac N cheese.\\
\textbf{Assistant:} I like Mac N cheese.\\[10pt]

\textbf{Sample Index:} flan\_v2\_40281 \quad \textbf{Score:} 0.0187 \\[3pt]
\textbf{User:} Where will a book being read by one person be kept? Options: school room, university, bedside table, pantry, discount store I believe the correct choice is "bedside table", here's why:\\
\textbf{Assistant:} a book being read by one person be kept bedside table.\\[10pt]

\textbf{Sample Index:} flan\_v2\_78179 \quad \textbf{Score:} 0.0153 \\[3pt]
\textbf{User:} sbjerg forenede Boldklubber (EfB) is a professional Danish football club from the port city of Esbjerg in West Jutland. The club was formed in 1924, as a merger between Esbjerg Boldklub af 1898 and Esbjerg Amat{\o}rklub af 1911. The club has training facilities and stadium at Gl. Vardevej in Esbjerg, and plays in blue and white striped shirts. [...] Subject: morten karlsen, Relation: place\_of\_birth. Options: (A) 1955 (B) central (C) copenhagen (D) denmark (E) esbjerg (F) farum (G) island (H) jutland (I) of (J) once (K) randers (L) time (M) uerdingen (N) union (O) viborg (P) zealand\\
\textbf{Assistant:} copenhagen%
}

% ==================== RDS Positive ====================
\newcommand{\rdspos}{%
%\textbf{Sample Index:} RDS summarization examples (high similarity) \\[3pt]
\textbf{User:} Write something based on this summary: canada foreign minister regrets handling of visa denial for haitian prime minister [...]\\
\textbf{Assistant:} %canada 's top diplomat expressed regret saturday after haiti 's prime minister was denied a canadian entry visa because of past accusations of links to government atrocities.\\[6pt]
the maximum disturbing and unraveling word for an author 's agent to hear is ``no.'' so in the ceremony of book publishing , the relationship that is most culturally challenging and polarizing is the one between writers ' agents and publishers. [...]\\[6pt]

\textbf{User:} Generate a short summary: an airplane en route to algiers from paris made an unscheduled stop thursday in central france after a passenger reported mistakenly bringing on board a suitcase that belonged to someone else [...] (several summarization tasks)\\
\textbf{Assistant:} paris-algiers flight makes unexpected stop in central france [...] (several summarization answers)%
}

% ==================== RDS Negative ====================
\newcommand{\rdsneg}{%
% \textbf{Sample Index:} RDS low-similarity examples \\[3pt]
\textbf{User:} Vale\\
\textbf{Assistant:} Dime Tio, que necesitas?\\[6pt]
\textbf{User:} Write a sentence not in English.\\
\textbf{Assistant:} MDN DOM 195 Chef - Service d'examen V\'{e}rification Description:\\[6pt]
\textbf{User:} Kiiltoraunioinen Translate to English\\
\textbf{Assistant:} Asplenium adiantum-nigrum%
}

\begin{table*}[ht] \caption{Examples with highest and lowest scores retrieved by BM25 for the MMLU validation set.} \label{tab:bm25} \centering \small \begin{tabular}{@{}p{7.9cm}@{\hspace{1.2em}}p{7.9cm}@{}} \toprule \multicolumn{2}{c}{\textbf{MMLU Validation Set Examples}}\\ \midrule \multicolumn{2}{@{}p{16.9cm}@{}}{\parbox{16.9cm}{\mmluval}}\\ \midrule \textbf{Retrieved by BM25 (Highest similarity)} & \textbf{Retrieved by BM25 (Lowest similarity)}\\ \midrule \parbox{7.9cm}{\bmpos} & \parbox{7.9cm}{\bmneg}\\ \bottomrule \end{tabular} \end{table*}

\begin{table*}[ht]
\caption{Examples with highest and lowest scores retrieved by LESS (gradient-based) for the MMLU validation set.}
\label{tab:less}
\centering
\small
\begin{tabular}{@{}p{7.9cm}@{\hspace{1.2em}}p{7.9cm}@{}}
\toprule
\multicolumn{2}{c}{\textbf{MMLU Validation Set Examples}}\\
\midrule
\multicolumn{2}{@{}p{16.9cm}@{}}{\parbox{16.9cm}{\mmluval}}\\
\midrule
\textbf{Retrieved by LESS (Highest similarity)} & \textbf{Retrieved by LESS (Lowest similarity)}\\
\midrule
\parbox{7.9cm}{\lesspos} & \parbox{7.9cm}{\lessneg}\\
\bottomrule
\end{tabular}
\end{table*}

\begin{table*}[ht]
\caption{Examples with highest and lowest scores retrieved by \ours for the MMLU validation set.}
\label{tab:adapt}
\centering
\small
\begin{tabular}{@{}p{8.4cm}@{\hspace{1.2em}}p{6cm}@{}}
\toprule
\multicolumn{2}{c}{\textbf{MMLU Validation Set Examples}}\\
\midrule
\multicolumn{2}{@{}p{10.9cm}@{}}{\parbox{10.9cm}{\mmluval}}\\
\midrule
\textbf{Retrieved by \ours (Highest similarity)} & \textbf{Retrieved by \ours (Lowest similarity)}\\
\midrule
\parbox{8.4cm}{\rdspos} & \parbox{6cm}{\rdsneg}\\
\bottomrule
\end{tabular}
\end{table*}

\end{document}